\documentclass[10pt,journal,compsoc]{IEEEtran}

\usepackage[breaklinks=true,letterpaper=true,colorlinks,bookmarks=false]{hyperref}

\usepackage{xspace}
\usepackage{graphicx}
\usepackage{booktabs}
\usepackage{multirow}
\usepackage{color}
\usepackage{colortbl}
\usepackage{amssymb}
\usepackage{subcaption}
\usepackage{cite}

\definecolor{mygray}{gray}{.9}

\usepackage{array}
\newcolumntype{x}[1]{>{\centering\arraybackslash\hspace{0pt}}p{#1}}

\newcommand{\etal}[0]{\textit{et al.\xspace}}
\newcommand{\eg}[0]{\emph{e.g.\xspace}}
\newcommand{\pb}[0]{\parbox{1.9cm}}

\ifCLASSINFOpdf
\else
\fi

\newcommand{\mh}[1]{{\textcolor{black}{#1}}}

\newcommand{\sk}[1]{{\textcolor{black}{#1}}}

\begin{document}

\title{Transformers in Vision: A Survey}

\author{Salman~Khan,~
        Muzammal~Naseer,~
        Munawar~Hayat,~%
        Syed~Waqas~Zamir,\\~%
        Fahad~Shahbaz~Khan,~%
        and~Mubarak~Shah
\IEEEcompsocitemizethanks{\IEEEcompsocthanksitem S. Khan, M. Naseer and F. S. Khan are with the MBZ University of Artificial Intelligence, Abu Dhabi, UAE.\protect\\
E-mail: firstname.lastname@mbzuai.ac.ae
\IEEEcompsocthanksitem M. Hayat is with the Faculty of IT, Monash University, Clayton VIC 3800, Australia.\protect
\IEEEcompsocthanksitem S. W. Zamir is with the Inception Institute of Artificial Intelligence, Abu Dhabi, UAE.\protect
\IEEEcompsocthanksitem S. Khan and M. Naseer are also with the CECS, Australian National University, Canberra ACT 0200, Australia. \protect
\IEEEcompsocthanksitem F. S. Khan is also with the Computer Vision Laboratory, Linköping University, Sweden. \protect
\IEEEcompsocthanksitem M. Shah is with the 
Center for Research in Computer Vision, University of Central Florida, Orlando, FL 32816, United States.
}
\thanks{Manuscript received March, 2021.}}

\markboth{ }%
{Shell \MakeLowercase{\textit{et al.}}: Bare Demo of IEEEtran.cls for Computer Society Journals}

\IEEEtitleabstractindextext{%
\begin{abstract}
Astounding results from Transformer models on natural language tasks have intrigued the vision community to study their application to computer vision problems. Among their salient benefits, Transformers enable modeling long dependencies between input sequence elements and support parallel processing of sequence as compared to recurrent networks \eg, Long short-term memory (LSTM). Different from convolutional networks, Transformers require minimal inductive biases for their design and are naturally suited as set-functions. Furthermore, the straightforward design of Transformers allows processing multiple modalities (\eg, images, videos, text and speech) using similar processing blocks and demonstrates excellent scalability to very large capacity networks and huge datasets. These strengths have led to exciting progress on a number of vision tasks using Transformer networks. This survey aims to provide a comprehensive overview of the Transformer models in the computer vision discipline. We start with an introduction to fundamental concepts behind the success of Transformers i.e., self-attention, large-scale pre-training, and bidirectional feature encoding. We then cover extensive applications of transformers in vision including popular recognition tasks (\eg, image classification, object detection, action recognition, and  segmentation), generative modeling, multi-modal tasks (\eg, visual-question answering, visual reasoning, and visual grounding), video processing (\eg, activity recognition, video forecasting), low-level vision (\eg, image super-resolution, image enhancement, and colorization) and 3D analysis (\eg, point cloud classification and segmentation). We compare the respective advantages and limitations of popular techniques both in terms of architectural design and their experimental value. Finally, we provide an analysis on open research directions and possible future works.  We hope this effort will ignite further interest in the community to solve current challenges towards the application of transformer models in computer vision. 
\end{abstract}

}

\maketitle

\IEEEdisplaynontitleabstractindextext

\IEEEpeerreviewmaketitle

\IEEEraisesectionheading{\section{Introduction}\label{sec:introduction}}

\IEEEPARstart{T}{ransformer}  
 models~\cite{vaswani2017attention} have recently demonstrated exemplary performance on a broad range of language tasks \eg, text classification, machine translation \cite{ott2018scaling} and question answering. Among these models, the most popular ones include BERT (Bidirectional Encoder Representations from Transformers)  \cite{devlin2018bert}, GPT (Generative Pre-trained Transformer) v1-3 \cite{radford2018improving,radford2019language,brown2020language}, RoBERTa (Robustly Optimized BERT Pre-training) \cite{liu2019roberta} and T5 (Text-to-Text Transfer Transformer) \cite{raffel2019exploring}. The profound impact of Transformer models has become more clear with their scalability to very large capacity models \cite{lepikhin2020gshard,fedus2021switch}. For example, the BERT-large \cite{devlin2018bert} model with 340 million parameters was significantly outperformed by the  GPT-3~\cite{brown2020language} model with 175 billion parameters while the latest mixture-of-experts Switch transformer \cite{fedus2021switch} scales up to a whopping 1.6 trillion parameters!
 
 \begin{figure*}[!t]
    \centering
    \includegraphics[width=\textwidth]{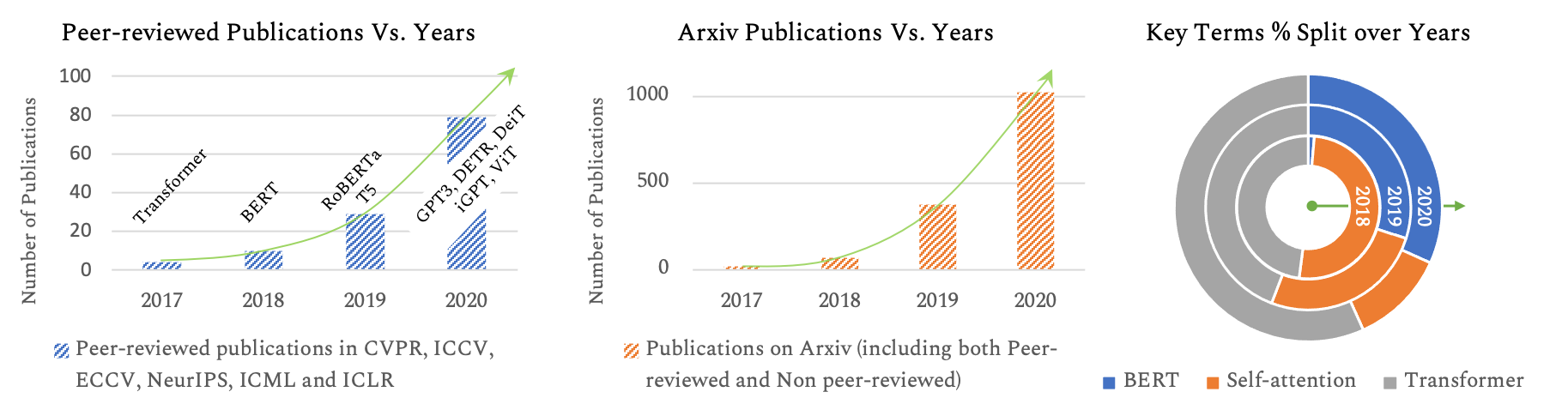}
    \vspace{-0.7cm}
    \caption{\small Statistics on the number of times keywords such as BERT, Self-Attention, and Transformers appear in the titles of Peer-reviewed and arXiv papers over the past few years (in Computer Vision and Machine Learning). The plots show consistent growth in recent literature. This survey covers recent progress on Transformers in the computer vision domain.}
    \label{fig:growth}
\end{figure*}

The breakthroughs from Transformer networks in Natural Language Processing (NLP) domain has sparked great interest in the computer vision community to adapt these models for vision and multi-modal learning tasks (Fig.~\ref{fig:growth}). However, visual data follows a typical structure (e.g., spatial and temporal coherence), thus demanding novel network designs and training schemes. As a result, Transformer models and their variants have been successfully used for image recognition \cite{vision_transformer,touvron2020deit}, object detection \cite{carion2020end,zhu2020deformable}, segmentation \cite{ye2019cross}, image super-resolution \cite{yang2020superresolution}, video understanding \cite{sun2019videobert,girdhar2019video}, image generation \cite{chen2020ipt}, text-image synthesis \cite{Ramesh2021dalle} and visual question answering \cite{tan2019lxmert,su2019vl}, among several other use cases \cite{wang2020sceneformer,anonymous2021colorization,doersch2020crosstransformers,ye2020few}. This survey aims to cover such recent and exciting efforts in the computer vision domain, providing a comprehensive reference to interested readers. 

Transformer architectures are based on a self-attention mechanism that learns the relationships between elements of a sequence. As opposed to recurrent networks that process sequence elements recursively and can only attend to short-term context, Transformers can attend to complete sequences thereby learning long-range relationships. 
\sk{Although attention models have been extensively used in both feed-forward and recurrent networks \cite{chaudhari2019attentive,de2021attention}, Transformers are based solely on the attention mechanism and have a unique implementation (i.e., multi-head attention) optimized for parallelization.} An important feature of these models is their scalability to high-complexity models and large-scale datasets \sk{e.g., in comparison to some of the other alternatives such as hard attention \cite{vinyals2015show} which is stochastic in nature and requires Monte Carlo sampling for sampling attention locations.} 
Since Transformers assume minimal prior knowledge about the structure of the problem as compared to their convolutional and recurrent counterparts~\cite{bengio2017deep,lecun2015deep,hochreiter1997long},  they are typically pre-trained using pretext tasks on large-scale (unlabelled) datasets \cite{vaswani2017attention,devlin2018bert}.  Such a pre-training avoids costly manual annotations, thereby encoding highly expressive and generalizable representations that model rich relationships between the entities present in a given dataset.  The learned representations are then fine-tuned on the downstream tasks in a supervised manner to obtain favorable results.



This paper provides a holistic overview of the transformer models developed for computer vision applications. We develop a taxonomy of the network design space and highlight the major strengths and shortcomings of the existing methods. Other literature reviews mainly focus on the NLP domain \cite{hu2019introductory,tay2020efficient} or cover generic attention-based approaches \cite{hu2019introductory,chaudhari2019attentive}. By focusing on the newly emerging area of visual transformers, we comprehensively organize the recent approaches according to the intrinsic features of self-attention and the investigated task. We first provide an introduction to the salient concepts underlying Transformer networks and then elaborate on the specifics of recent vision transformers. 
Where ever possible, we draw parallels between the Transformers used in the NLP domain \cite{vaswani2017attention} and the ones developed for vision problems to flash major novelties and interesting domain-specific insights.  
Recent approaches show that convolution operations can be fully replaced with attention-based transformer modules and have also been used jointly in a single design to encourage symbiosis between the two complementary set of operations. 
This survey finally details open research questions with an outlook towards the possible future work.  

\section{Foundations}
There exist two key ideas that have contributed towards the development of conventional transformer models. 
(a) The first one is \emph{self-attention}, which allows capturing `long-term' dependencies between sequence elements as compared to conventional recurrent models that find it challenging to encode such relationships.  
(b) The second key idea is that of \emph{pre-training}\footnote{\sk{Several recent Vision Transformers demonstrate that the model can be learned end-to-end on ImageNet-1K without any dedicated pre-training phase \cite{yuan2021tokens,liu2021swin,chu2021twins}. 
However, the performance generally remains lower than the pre-trained counter-parts.}} on a large (un)labelled corpus in a \sk{(self)supervised manner}, and subsequently fine-tuning to the target task with a small labeled dataset \cite{devlin2018bert,liu2019roberta,wang2020linformer}. 
Below, we provide a brief tutorial on these two ideas (Sec.~\ref{sec: Self-supervision} and~\ref{sec: Self-Attention}), along with a summary of seminal Transformer networks (Sec.~\ref{sec: Transformer Model} and~\ref{sec: The Bidirectional Representations}) where these ideas have been applied. 
This background will help us better understand the forthcoming Transformer based models used in the computer vision domain (Sec.~\ref{sec:transformers_vision}).

\begin{figure}[t]
    \centering
    \includegraphics[width=1\columnwidth]{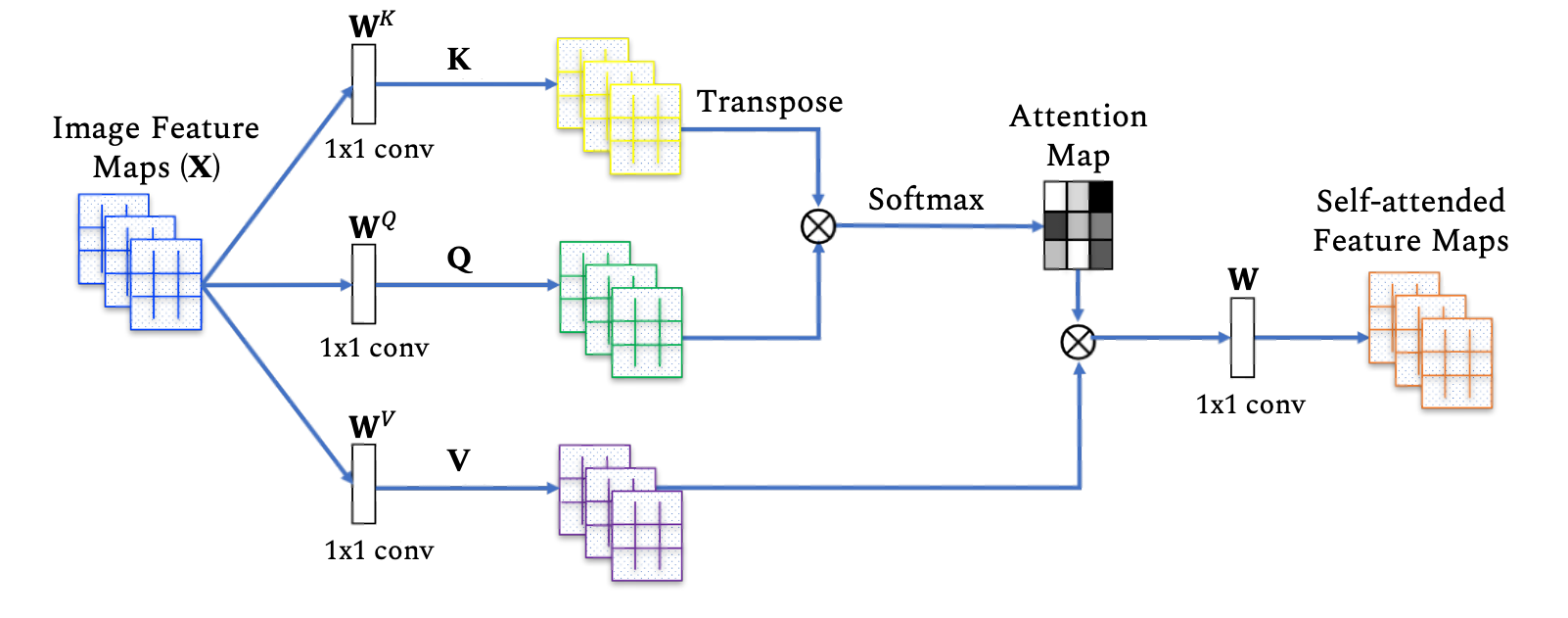}
    \caption{\small 
    An example self-attention block used in the vision domain \cite{zhang2019self}. 
    Given the input sequence of image features, the triplet of (key, query, value) is calculated followed by attention calculation and applying it to reweight the values. 
    A single head is shown here and an output projection ($\textbf{W}$) is finally applied to obtain output features with the same dimension as the input. Figure adapted from \cite{zhang2019self}.}
    \vspace{-0.4cm}
    \label{fig:self-attention}
\end{figure}

 \begin{figure*}[t]
     \centering
     \includegraphics[width=0.99\textwidth]{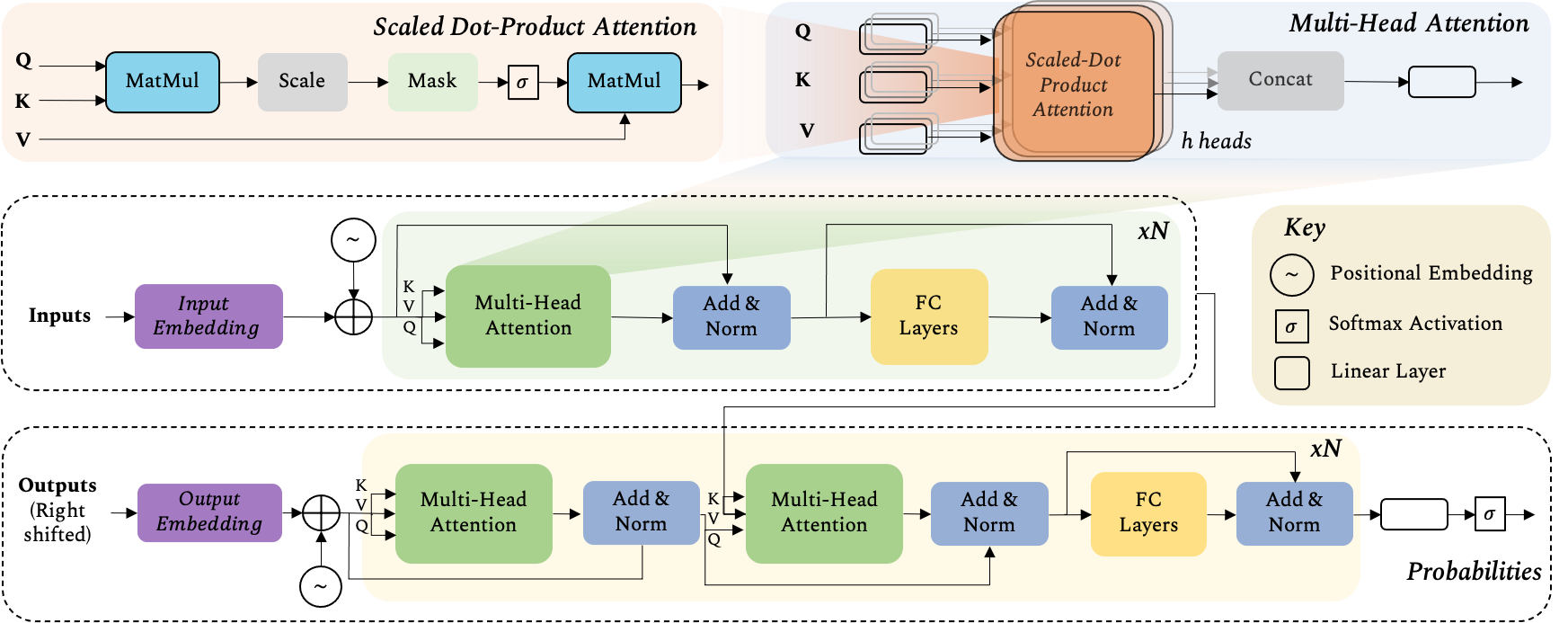}
     \caption{\small \emph{Architecture of the Transformer Model} \cite{vaswani2017attention}. The model was first developed for the language translation task where an input sequence in one language is required to be converted to the output sequence in another language. The Transformer encoder (\emph{middle} row) operates on the input language sequence and converts it to an embedding before passing it on to the encoder blocks. 
     The Transformer decoder (\emph{bottom} row) operates on the previously generated outputs in the translated language and the encoded input sequence from the middle branch to output the next word in the output sequence. The sequence of previous outputs (used as input to the decoder) is obtained by shifting the output sentence to the right by one position and appending start-of-sentence token at the beginning. This shifting avoids the model to learn to simply copy the decoder input to the output. The ground-truth to train the model is simply the output language sequence (without any right shift) appended with an end-of-sentence token. The blocks consisting of multi-head attention (\emph{top} row) and feed-forward layers are repeated $N$ times in both the encoder and decoder.
     }
     \label{fig:transformer}
 \end{figure*}

\subsection{\sk{Self-Attention in Transformers}}
\label{sec: Self-Attention}
Given a sequence of items, self-attention estimates the relevance of one item to other items (e.g., which words are likely to come together in a sentence). The self-attention mechanism is an integral component of Transformers, which explicitly models the interactions between all entities of a sequence for structured prediction tasks. Basically, a self-attention layer updates each component of a sequence by aggregating global information from the complete input sequence. 
Lets denote a sequence of $n$ entities ($\mathbf{x}_1, \mathbf{x}_2, \cdots \mathbf{x}_n$) by $\mathbf{X} \in \mathbb{R}^{n \times d}$, where $d$ is the embedding dimension to represent each entity. The goal of self-attention is to capture the interaction amongst all $n$ entities by encoding each entity in terms of the global contextual information. This is done by defining three learnable weight matrices to transform Queries ($\mathbf{W}^Q \in \mathbb{R}^{d \times d_q}$), Keys ($\mathbf{W}^K \in \mathbb{R}^{d \times d_k}$) and Values ($\mathbf{W}^V \in \mathbb{R}^{d \times d_v}$), where $d_q=d_k$. The input sequence $\mathbf{X}$ is first projected onto these weight matrices to get $\mathbf{Q}=\mathbf{X}\mathbf{W}^Q$, $\mathbf{K}=\mathbf{X}\mathbf{W}^K$ and $\mathbf{V}=\mathbf{X}\mathbf{W}^V$. The output $\mathbf{Z}\in\mathbb{R}^{n \times d_v}$ of the self attention layer is,
$$\mathbf{Z}= \mathbf{softmax}\left (\frac{\mathbf{Q}\mathbf{K}^T}{\sqrt{d_q}}\right )\mathbf{V}.$$
For a given entity in the sequence, the self-attention basically computes the dot-product of the query with all keys, which is then normalized using softmax operator to get the attention scores. Each entity then becomes the weighted sum of all entities in the sequence, where weights are given by the attention scores (Fig.~\ref{fig:self-attention} and Fig.~\ref{fig:transformer}, top row-left block).

\textbf{Masked Self-Attention:} The standard self-attention layer attends to all entities. For the Transformer model \cite{vaswani2017attention} which is trained to predict the next entity of the sequence, the self-attention blocks used in the decoder are masked to prevent attending to the subsequent future entities. This is simply done by an element-wise multiplication operation with a mask $\mathbf{M} \in \mathbb{R}^{n \times n}$, where $\mathbf{M}$ is an upper-triangular matrix. The masked self-attention is defined by,  $$ \mathbf{softmax}\left (\frac{\mathbf{Q}\mathbf{K}^T}{\sqrt{d_q}} \circ \mathbf{M}\right ),$$ where $\circ$ denotes Hadamard product. Basically, while predicting an entity in the sequence, the attention scores of the future entities are set to zero in masked self-attention.

\textbf{Multi-Head Attention:}
In order to encapsulate multiple complex relationships amongst different elements in the sequence, the multi-head attention comprises multiple self-attention blocks ($h=8$ in the original Transformer model \cite{vaswani2017attention}). Each block has its own set of learnable weight matrices $\{\mathbf{W}^{Q_i},\mathbf{W}^{K_i},\mathbf{W}^{V_i} \}$, where $i=0 \cdots (h{-}1)$. For an input $\mathbf{X}$, the output of the $h$ self-attention blocks in multi-head attention is then concatenated into a single matrix $[\mathbf{Z}_0,\mathbf{Z}_1,\cdots\mathbf{Z}_{h-1}] \in \mathbb{R}^{n\times h \cdot d_v}$ and projected onto a weight matrix $\mathbf{W} \in \mathbb{R}^{h \cdot d_v \times d}$ (Fig.~\ref{fig:transformer}, top row).

The main difference of self-attention with convolution operation is that the filters are dynamically calculated instead of static filters (that stay the same for any input) as in the case of convolution. Further, self-attention is invariant to permutations and changes in the number of input points. \sk{As a result, it can easily operate on irregular inputs as opposed to standard convolution that requires grid structure. Furthermore, it has been shown in the literature how self-attention (with positional encodings) is theoretically a more flexible operation which can model the behaviour of convolutional models towards encoding local features \cite{perez2018turing}. Cordonnier \etal \cite{cordonnier2019relationship} further studied the relationships between self-attention and convolution operations. Their empirical results confirm that multi-head self-attention (with sufficient parameters) is a more generic operation which can model the expressiveness of convolution as a special case. In fact, self-attention provides the capability to learn the global as well as local features, and provide expressivity to adaptively learn kernel weights as well as the receptive field (similar to deformable convolutions \cite{dai2017deformable}).}

\subsection{\sk{(Self) Supervised Pre-training}}
\label{sec: Self-supervision}
Self-attention based Transformer models generally operate in a two-stage training mechanism. First, pre-training is performed on a large-scale dataset (and sometimes a combination of several available datasets \cite{su2019vl,chen2020uniter}) in either a supervised \cite{vision_transformer}  or a \sk{self-supervised manner} \cite{devlin2018bert,li2020oscar,lin2020end}. Later, the pre-trained weights are adapted to the down-stream tasks using small-mid scale datasets. Examples of downstream tasks include image classification \cite{gidaris2018unsupervised}, object detection \cite{carion2020end}, \sk{zero-shot classification} \cite{Ramesh2021dalle}, question-answering \cite{fedus2021switch} and action recognition \cite{girdhar2019video}. The effectiveness of pre-training for large-scale Transformers has been advocated in both the language and vision domains. For example, Vision Transformer model (ViT-L) \cite{vision_transformer} experiences an absolute $13\%$ drop in accuracy on ImageNet test set when trained only on ImageNet train set as compared to the case when pretrained on JFT dataset \cite{JFT300M} with 300 million images.

Since acquiring manual labels at a massive scale is cumbersome, self-supervised learning has been very effectively used in the pre-training stage.  The self-supervision based pre-training stage training has played a crucial role in unleashing the scalability and generalization of Transformer networks, enabling training even above a \emph{trillion} parameter networks (e.g., the latest Switch Transformer \cite{fedus2021switch} from Google). 
An extensive survey on SSL can be found in \cite{jing2020self,liu2020self}. As nicely summarized by Y.~LeCun \cite{lecun_keynote}, the basic idea of SSL is to \emph{fill in the blanks}, i.e., try to predict the occluded data in images, future or past frames in temporal video sequences or predict a pretext task \eg, the amount of rotation applied to inputs, the permutation applied to image patches or the color of a gray-scale image. Another effective way to impose self-supervised constraints is via contrastive learning. In this case, nuisance transformations are used to create two types of modified versions of the same image i.e.,  without changing the underlying class semantics (\eg, image stylizing, cropping) and with semantic changes (\eg, replacing an object with another in the same scene, or changing the class with minor adversarial changes to the image). Subsequently, the model is trained to be invariant to the nuisance transformations and emphasize on modeling minor changes that can alter semantic labels. 

Self-supervised learning provides a promising learning paradigm since it enables learning from a vast amount of readily available non-annotated data. In the SSL based pre-training stage, a model is trained to learn a meaningful representation of the underlying data by solving a pretext task. The pseudo-labels for the pretext task are automatically generated (without requiring any expensive manual annotations) based on data attributes and task definition. Therefore, the pretext task definition is a critical choice in SSL. We can broadly categorize existing SSL methods based upon their pretext tasks into \textbf{(a)} \textit{generative} approaches which synthesize images or videos (given conditional inputs), \textbf{(b)} \textit{context-based} methods which exploit the relationships between image patches or video frames, and \textbf{(c)} \textit{cross-modal} methods which leverage from multiple data modalities. Examples of \textit{generative} approaches include conditional generation tasks such as masked image modeling \cite{chen2020uniter} and image colorization \cite{zhang2016colorful}, image super-resolution \cite{ledig2017photo}, image in-painting \cite{pathak2016context}, and GANs based methods \cite{Goodfellow2014, lin2017marta}. The \textit{context-based} pretext methods solve problems such as a jigsaw puzzle on image patches \cite{ahsan2019video, noroozi2016unsupervised, Kim_2018}, masked object classification \cite{su2019vl}, predict geometric transformation such as rotation \cite{gidaris2018unsupervised, jing2018self}, or verify temporal sequence of video frames \cite{lee2017unsupervised,misra2016shuffle,wei2018learning}. Cross-modal pretext methods verify the correspondence of two input modalities \eg, text \& image \cite{li2019visualbert}, audio \& video \cite{korbar2018cooperative,arandjelovic2017look} or RGB \& flow \cite{sayed2018cross}.

\subsection{Transformer Model}\label{sec: Transformer Model}
The architecture of the Transformer model proposed in \cite{vaswani2017attention} is shown in Fig.~\ref{fig:transformer}. It has an encoder-decoder structure. The encoder (\emph{middle} row) consists of six identical blocks (i.e., $N{=}6$ in Fig.~\ref{fig:transformer}), with each block having two sub-layers: a multi-head self-attention network, and a simple position-wise fully connected feed-forward network. Residual connections \cite{he2016deep} alongside layer normalization \cite{ba2016layer} are employed after each block as in Fig.~\ref{fig:transformer}. Note that, different from regular convolutional networks where feature aggregation and feature transformation are simultaneously performed (\eg, with a convolution layer followed by a non-linearity), these two steps are decoupled in the Transformer model i.e., self-attention layer only performs aggregation while the feed-forward layer performs transformation.
Similar to the encoder, the decoder (\emph{bottom} row) in the Transformer model comprises six identical blocks. Each decoder block has three sub-layers, first two (multi-head self-attention, and feed-forward) are similar to the encoder, while the third sub-layer performs multi-head attention on the outputs of the corresponding encoder block, as shown in Fig.~\ref{fig:transformer}.

The original Transformer model in \cite{vaswani2017attention} was trained for the Machine Translation task. The input to the encoder is a sequence of words (sentence) in one language. \textbf{Positional encodings} are added to the input sequence to capture the relative position of each word in the sequence. Positional encodings have the same dimensions as the input $d=512$, and can be learned or pre-defined \eg, by sine or cosine functions. Being an auto-regressive model, the decoder of the Transformer \cite{vaswani2017attention} uses previous predictions to output the next word in the sequence. The decoder, therefore, takes inputs from the encoder as well as the previous outputs to predict the next word of the sentence in the translated language. To facilitate residual connections the output dimensions of all layers are kept the same i.e., $d=512$. The dimensions of query, key and value weight matrices in multi-head attention are set to $d_q=64, d_k=64, d_v=64$.

\begin{figure*}[htp]
     \centering
     \includegraphics[trim=0mm 4.5cm 0mm 0mm, clip,width=0.85\textwidth]{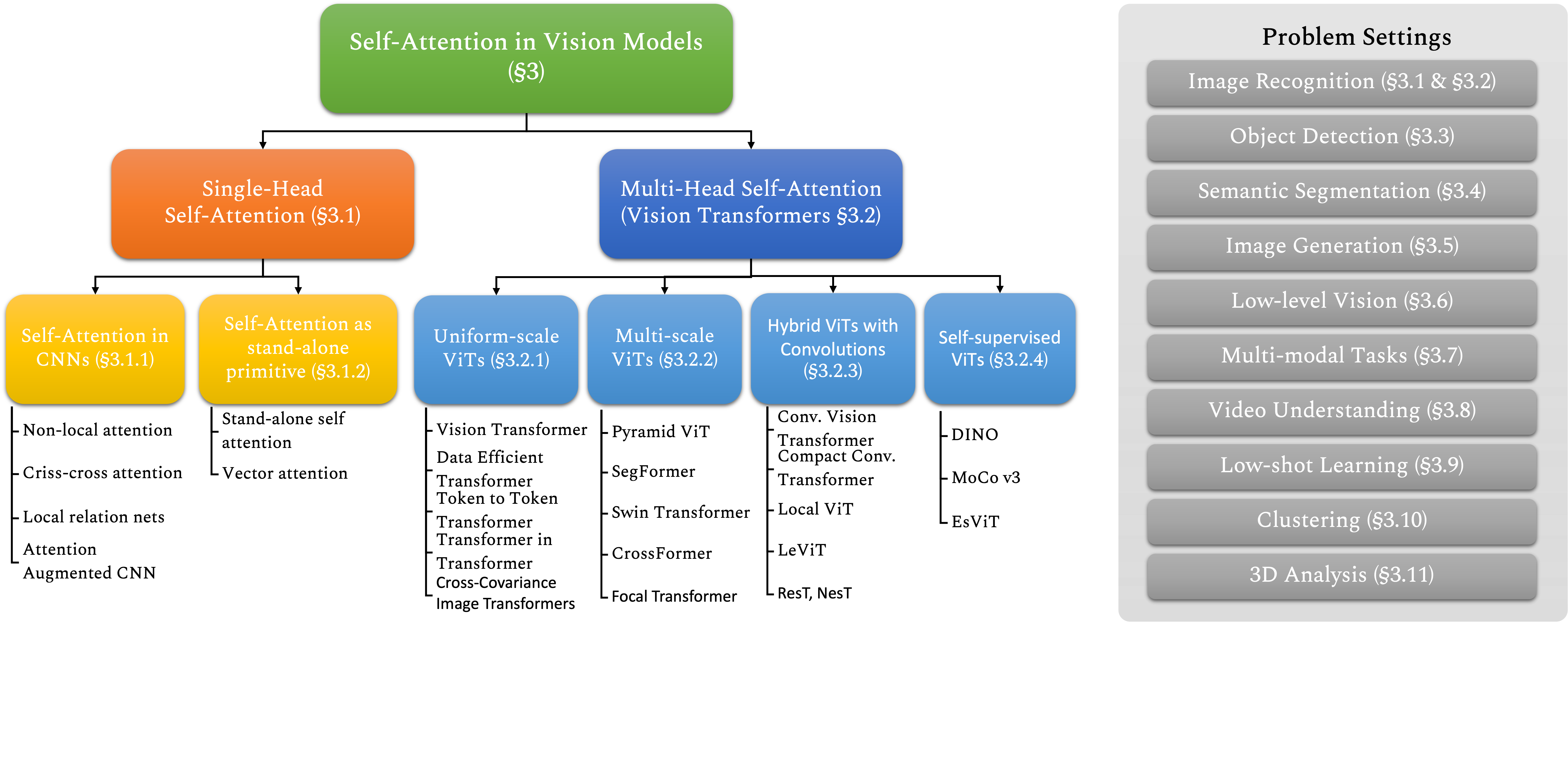}
     \caption{\small \sk{\emph{A taxonomy of self-attention design space}. Existing approaches based on self-attention explore single-head or multi-head (transformer) designs for vision tasks. We note that interesting efforts have been made to utilize knowledge from convolution based architectures to improve ViTs (e.g., multi-scale and hybrid designs). We categorize the upcoming sections of this survey according to the types of self-attention block (\emph{left tree diagram}) as well as the prominent tasks in computer vision (\emph{right}). }}
     \vspace{-0.5cm}
     \label{fig:taxonomy_of_attention}
 \end{figure*}

\subsection{Bidirectional Representations} \label{sec: The Bidirectional Representations}

The training strategy of the original Transformer model \cite{vaswani2017attention} could only attend to the context on the left of a given word in the sentence. This is limiting, since for most language tasks, contextual information from both left and right sides is important. Bidirectional Encoder Representations from
Transformers (BERT) \cite{devlin2018bert} proposed to jointly encode the right and left context of a word in a sentence, thus improving the learned feature representations for textual data in an self-supervised manner. To this end, BERT \cite{devlin2018bert} introduced two pretext tasks to pre-train the Transformer model \cite{vaswani2017attention} in a self-supervised manner: \textit{Masked Language Model} and \textit{Next Sentence Prediction}. For adapting the pre-trained model for downstream tasks, a task-specific additional output module is appended to the pre-trained model, and the full model is fine-tuned end-to-end. 
Here, we briefly touch upon the pretext tasks. \textbf{(1) Masked Language Model (MLM) -} A fixed percentage (15\%) of words in a sentence are randomly masked and the model is trained to predict these masked words using cross-entropy loss. In predicting the masked words, the model learns to incorporate the bidirectional context. 
\textbf{(2) Next Sentence Prediction (NSP) -} Given a pair of sentences, the model predicts a binary label i.e., whether the pair is valid from the original document or not. The training data for this can easily be generated from any monolingual text corpus. A pair of sentences \textit{A} and \textit{B} is formed, such that \textit{B} is the actual sentence (next to \textit{A}) 50\% of the time, and \textit{B} is a random sentence for other 50\% of the time.  NSP enables the model to capture sentence-to-sentence relationships which are crucial in many language modeling tasks such as Question Answering and Natural Language Inference.

\section{Self-Attention \& Transformers in Vision}\label{sec:transformers_vision}


\mh{We broadly categorize vision models with self-attention into two categories: the models which use single-head self-attention (Sec.~\ref{Single-head Self-Attention}), and the models which employ multi-head self-attention based Transformer modules into their architectures (Sec.~\ref{Multi-head Self-Attention (Transformers)}). Below, we first discuss the first category of single-head self-attention based frameworks, which generally apply global or local self-attention within CNN architectures, or utilize matrix factorization to enhance design efficiency and use vectorized attention models. We then discuss the Transformer-based vision architectures in Sec.~\ref{Multi-head Self-Attention (Transformers)}.}


\subsection{\sk{Single-head Self-Attention}}\label{Single-head Self-Attention}

\subsubsection{Self-Attention in CNNs}

Inspired by non-local means operation \cite{buades2005non} which was mainly designed for image denoising, 
Wang \etal \cite{wang2018non} proposed a differentiable non-local operation for deep neural networks to capture long-range dependencies both in space and time in a feed-forward fashion. Given a feature map, their proposed operator \cite{wang2018non} computes the response at a position as a weighted sum of the features at all positions in the feature map. This way, the non-local operation is able to capture interactions between any two positions in the feature map regardless of the distance between them. Videos classification is an example of a task where long-range interactions between pixels exist both in space and time. Equipped with the capability to model long-range interactions, \cite{wang2018non} demonstrated the superiority of non-local deep neural networks for more accurate video classification on Kinetics dataset \cite{kay2017kinetics}.

\begin{figure}[tp]
\centering
    \begin{subfigure}[t]{0.48\columnwidth}
      \includegraphics[width=\textwidth]{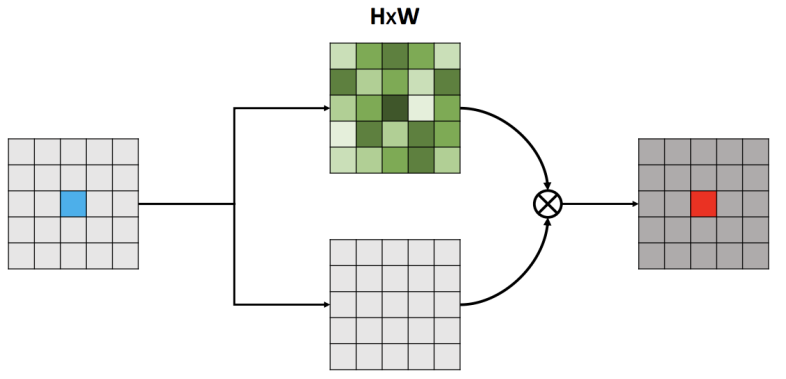}
      \caption{\small Non-local block~\cite{wang2018non}}
      \label{}
    \end{subfigure}\hspace{0.03cm}
    \begin{subfigure}[t]{0.48\columnwidth}
      \includegraphics[width=\textwidth]{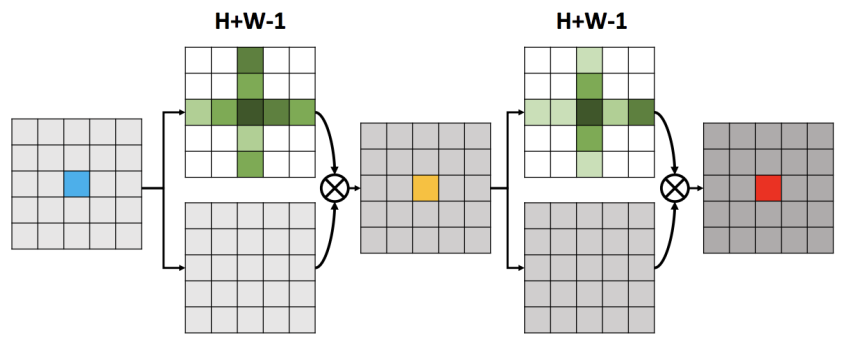}
      \caption{\small Criss-cross attention~\cite{huang2019ccnet} }
      \label{fig:dalle img_comp}
    \end{subfigure}
\caption{\small Comparison of two different self-attention approaches: Non-local self-attention block~\cite{wang2018non} and Criss-cross self-attention module \cite{huang2019ccnet}. Figure is from \cite{huang2019ccnet}.}
\label{fig:ccnet}
\end{figure}

Although the self-attention allows us to model full-image contextual information, it is both memory and compute intensive. As shown in Fig.~\ref{fig:ccnet}(a), in order to encode global context for a given pixel location, non-local block~\cite{wang2018non} computes a \emph{dense} attention map (in green). The non-local block~\cite{wang2018non} has a high complexity of $\mathcal{O}(N^2)$, where $N$ denotes the number of input feature maps. 
To reduce this computational burden, Huang \etal \cite{huang2019ccnet} propose the criss-cross attention module that for each pixel position generates a \emph{sparse} attention map only on the criss-cross path, as illustrated in  Fig.~\ref{fig:ccnet}(b). Further, by applying criss-cross attention recurrently, each pixel position can capture context from all other pixels. Compared to non-local block, the criss-cross uses 11$\times$ lesser GPU memory, and has a complexity of $\mathcal{O}(2\sqrt{N})$. 
State-of-the-art results are reported \cite{huang2019ccnet} for the semantic and instance segmentation tasks on several benchmark datasets including Cityscapes~\cite{cordts2016cityscapes}, ADE20K~\cite{zhou2017ade20k}, COCO~\cite{lin2014coco}, LIP~\cite{liang2018LIP} and CamVid~\cite{brostow2009camvid}.

Another shortcoming of the convolutional operator comes from the fact that after training, it applies fixed weights regardless of any changes to the visual input. Hu \etal \cite{hu2019local} proposed local relation networks to adaptively compose pixels in a local window. They introduced a new differentiable layer 
that adapts its weight aggregation based on the compositional relations (similarity) between pixels/features within a local window. Such adaptive weight aggregation introduces geometric priors into the network which are important for the recognition tasks \cite{hu2019local}. Convolution is considered to be a top-down operator as it remains fixed across positions while a non-local operation such as introduced in \cite{buades2005non} is a bottom-up method as it aggregates input features over the full image. The local relation layer belongs to the category of bottom-up methods but it is restricted to a fixed window size  \eg, 7x7 neighborhood.


Bello \etal~\cite{bello2019attention} explore the possibility of employing self-attention as an alternative to convolutional operators. They employ the relative position encoding~\cite{shaw2018self} in two dimensions to develop a new self-attention mechanism that maintains translation equivariance, a desirable property for handling images. Although this self-attention provides competitive results as a stand-alone computational primitive, the best performance is obtained in combination with the convolutional operations. Authors show that attention augmentation leads to systematic performance gains in image classification and object detection for different architectures.

\subsubsection{Self-Attention as Stand-alone Primitive}

As discussed above, convolutional layers possess translation equivariance but can not scale with a large receptive field, therefore can not capture long-range interactions \cite{parmar2019stand}. On the other hand, global attention \cite{vaswani2017attention} which attend to all spatial locations of the input can be computationally intensive and is preferred on down-sampled small images,  image patches \cite{vision_transformer} or augmenting the convolutional features space \cite{bello2019attention}. Ramachandran \etal \cite{parmar2019stand}   proposed to replace convolutional layers in deep neural networks with a local self-attention layer which can be applied to small or large inputs without increasing the computational cost. At a basic level, the proposed self-attention layer \cite{parmar2019stand} considers all pixel positions in a specific window size around a given pixel, compute queries, keys and value vectors for these pixels, and then aggregates the spatial information within this window. The value vectors are aggregated after projecting the softmax score of queries and keys. This process is repeated for all given pixels and the response is concatenated to produce the output pixel. ResNet models with local self-attention layer can solve ImageNet and COCO object detection with fewer parameters as compared to ResNet models based on convolutional layers \cite{parmar2019stand}.

Zhao \etal \cite{zhao2020exploring} note that a traditional convolution operator performs feature aggregation and transformation jointly (by applying a filter and then passing it through a non-linearity). In contrast, they propose to perform feature aggregation separately with self-attention followed by transformation using an element-wise perceptron layer. 
For feature aggregation, they propose two alternate strategies: (a) pairwise self-attention and (b) patch-wise self-attention. The pairwise self-attention is permutation and cardinality invariant operation, while the patch-wise self-attention does not have such invariance properties (similar to convolution). Both pairwise and patch-wise self-attentions are implemented as a \emph{vector} attention \cite{zhao2020exploring} that learns weights for both the spatial and channel dimensions. This provides an alternate approach for attention that is conventionally performed using scalar weights (by taking a dot-product). The pairwise self-attention is a set operator that computes a \emph{vector attention} keeping in view the relationships of a particular feature with its neighbors in a given local neighborhood. In contrast, patch-wise self-attention is a generalization of the convolution operator (not a set operator) and looks at all the feature vectors in the local neighbourhood when deriving the attention vectors. Authors show that with considerably fewer parameters, self-attention networks (SAN) can beat ResNet baselines on the ImageNet dataset. They further show robustness against adversarial perturbations \cite{szegedy2013intriguing,naseer2019cross} and generalization to unseen transformations \cite{naseer2021improving}. This behaviour is due to the dynamic nature of attention that makes it difficult for the adversary to calculate useful fooling directions.

\subsection{\sk{Multi-head Self-Attention (Transformers)}}\label{Multi-head Self-Attention (Transformers)}
\mh{Unlike the approaches discussed in Sec.~\ref{Single-head Self-Attention} which insert self-attention as a component in CNN inspired architectures, Vision Transformer (ViTs) \cite{vision_transformer} adapts the architecture of \cite{vaswani2017attention} (see Fig.~\ref{fig:transformer}), which cascades multiple Transformer layers. ViTs have gained significant research attention, and a number of recent approaches have been proposed which build upon ViTs. Below, we discuss these methods by categorizing them into: uniform scale ViTs having single-scale features through all layers (Sec.~\ref{Uniform-scale Vision Transformers}), multi-scale ViTs that learn hierarchical features which are more suitable for dense prediction tasks (Sec.~\ref{Hierarchical Multi-Stage ViTs for Dense Prediction}), and hybrid designs having convolution operations within ViTs (Sec.~\ref{Hybrid ViTs with Convolutions}).}

\subsubsection{\sk{Uniform-scale Vision Transformers}} \label{Uniform-scale Vision Transformers}
\sk{The original Vision Transformer \cite{vision_transformer} model belongs to this family, where the multi-head self-attention is applied to a consistent scale in the input image where the spatial scale is maintained through the network hierarchy. We name such models as the uniform-scale ViTs, as described below.}

\begin{figure}[]
     \centering
     \includegraphics[width=0.95\columnwidth]{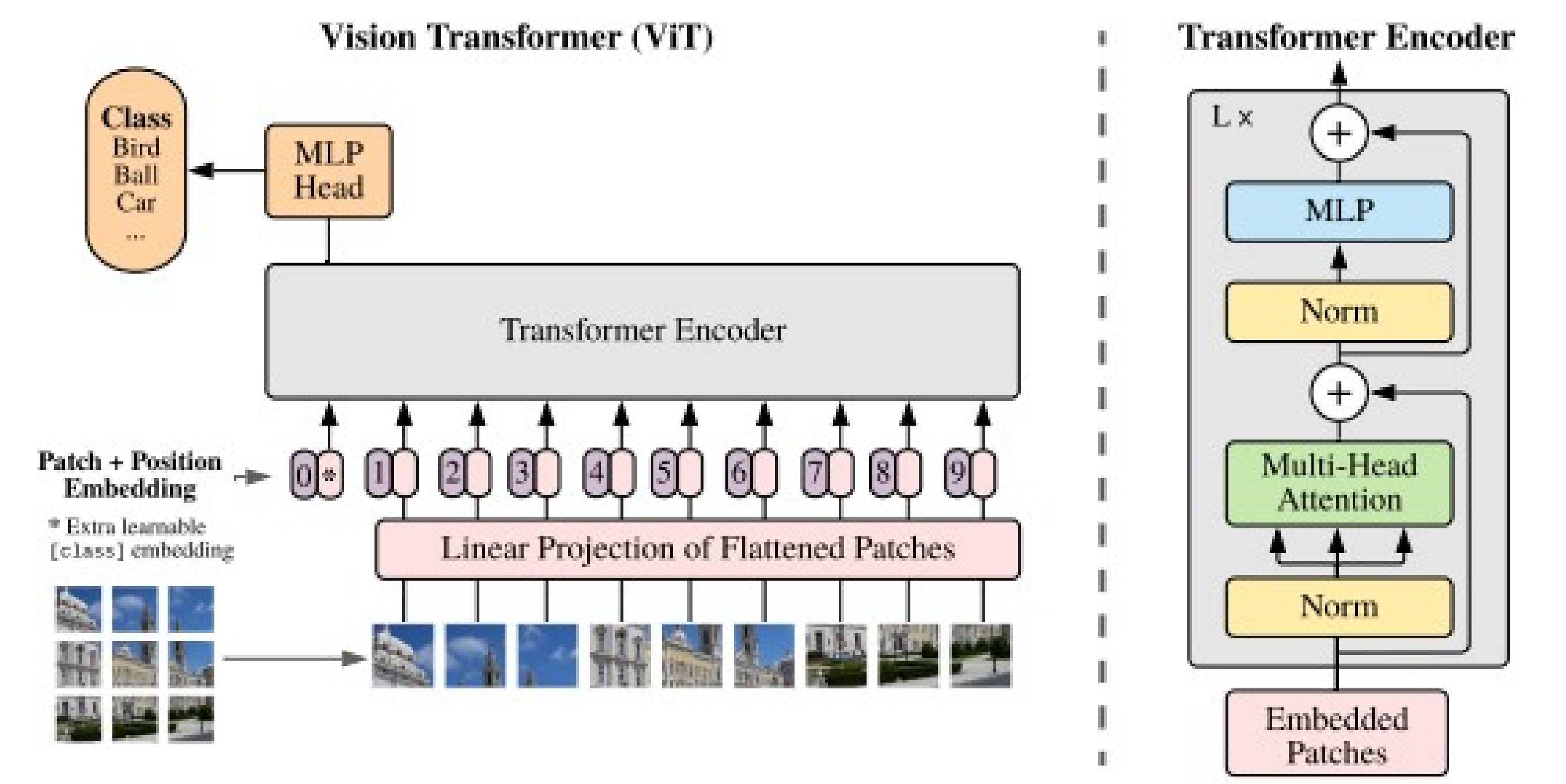}
     \caption{\small An overview of Vision Transformer (on the \emph{left}) and the details of Transformer encoder (on the \emph{right}). The architecture resembles Transformers used in the NLP domain and the image patches are simply fed to the model after flattening. After training, the feature obtained from the first token position is used for classification. Image obtained from \cite{vision_transformer}. }
     \label{fig:vision_transformer}
 \end{figure}

Vision Transformer (ViT) \cite{vision_transformer} (Fig.~\ref{fig:vision_transformer}) is the first work to showcase how Transformers can `altogether' replace standard convolutions in deep neural networks on large-scale image datasets. They applied the original Transformer model \cite{vaswani2017attention} (with minimal changes) on a sequence of image 'patches' flattend as vectors. The model was pre-trained on a large propriety dataset (JFT dataset \cite{JFT300M} with 300 million images) and then fine-tuned to downstream recognition benchmarks \eg, ImageNet classification. This is an important step since pre-training ViT on a medium-range dataset would not give competitive results, because the CNNs encode prior knowledge about the images (inductive biases \eg, translation equivariance) that reduces the need of data as compared to Transformers which must discover such information from very large-scale data. Notably, compared to the iGPT \cite{chen2020ipt} model that also applied Transformers to full-sized images but performs training as a generative task, ViT pre-trains the model with a supervised classification task (although a self-supervision variant is also explored which results in a less performance). 

The DeiT \cite{touvron2020deit} is the first work to demonstrate that Transformers can be learned on mid-sized datasets (i.e., 1.2 million ImageNet examples compared to 300 million images of JFT \cite{vision_transformer} used in ViT \cite{vision_transformer}) in relatively shorter training episodes. 
Besides using augmentation and regularization procedures common in CNNs, the main contribution of DeiT \cite{touvron2020deit} is a novel native distillation approach for Transformers which uses a CNN as a teacher model (RegNetY-16GF \cite{radosavovic2020designing}) to train the Transformer model. The outputs from the CNN aid the Transformer in efficiently figuring out useful representations for input images. A distillation token is appended with the input patch embeddings and the class token. The self-attention layers operate on these tokens to learn their inter-dependencies and outputs the learned class, patch, and distillation tokens. The network is trained with a cross-entropy loss defined on the output class token and a distillation loss to match the distillation token with the teacher output. Both \emph{soft} and \emph{hard} label choices were explored for distillation, where the hard distillation was found to perform better. Interestingly, the learned class and distillation tokens do not exhibit a high correlation indicating their complementary nature. The learned representations compare favorably well against top-performing CNN architectures such as EfficientNet \cite{tan2019efficientnet} and also generalize well for a number of downstream recognition tasks.

Token to Token (T2T) ViT \cite{yuan2021tokens} recursively combines neighboring tokens into a single token to reduce tokens length and aggregate spatial context. Transformer in Transformer \cite{han2021transformer} computes attention at two levels: patch-level (as done is standard ViTs \cite{vision_transformer}) and local sub-patch-level (\eg by subdividing a $16\times16$ patch into four $4\times4$ blocks, and computing attention amongst these blocks). 
In token labelling ViT \cite{jiang2021all}, all patch tokens contribute towards loss calculation, different from regular ViTs that only use classification token in the loss. This process includes auxiliary supervision where each image-patch (token) is labeled using a pre-trained CNN model. Similar to CutMix augmentation \cite{yun2019cutmix}, tokens from different images are mixed as an augmentation strategy, and the model is trained using the standard classification loss and auxiliary token-label loss. Their model demonstrates excellent performance specially for smaller sized models.

The quadratic complexity of self-attention hinders its applicability to longer sequences (high-resolution images). Cross-Covariance Image Transformers (XCiT) \cite{elnouby2021xcit} incorporate attention across feature-channels instead of tokens, i.e., their cross-covariance attention is given by $\mathbf{V} \mathbf{softmax}\left (\frac{\mathbf{K}^T \mathbf{Q}^T}{\sqrt{\tau}}\right )$. The proposed cross-covariance attention has linear complexity (since it depends upon feature dimension instead of the number of tokens). XCiT can therefore handle large resolution images and demonstrate excellent performance across different vision tasks i.e., self-supervised and fully supervised image classification and dense prediction (detection, segmentation). DeepViT \cite{zhou2021deepvit} observes that the similarity between attention maps of deeper layer is high and hinders scaling models depth. They propose to re-attend the attention maps in a multi-head block instead of simple aggregation of these attention maps, and show consistent gains over standard multi-head self attention based ViTs.

\subsubsection{Multi-scale Vision Transformers}
\label{Hierarchical Multi-Stage ViTs for Dense Prediction}
In standard ViTs, the number of the tokens and token feature dimension are kept fixed throughout different blocks of the network. This is limiting, since the model is unable to capture fine spatial details at different scales. Initial Transformer based dense prediction methods (e.g., DETR \cite{carion2020end}) therefore have a convolutional backend. Multi-stage hierarchical design for ViTs, where number of tokens is gradually reduced while the token feature dimension is  progressively increased, has been shown to produce effective features for dense prediction tasks \cite{wang2021pyramid, yang2021focal, liu2021swin, huang2021shuffle, wu2021cvt}. These models generally also perform well for recognition tasks. These architectures mostly sparsify tokens by merging neighboring tokens and projecting them to a higher dimensional feature space. Examples of multi-stage ViTs include Pyramid ViT \cite{wang2021pyramid, wang2021pvtv2}, Twins \cite{chu2021twins}, CoaT \cite{xu2021coscale}, Swin Transformer \cite{liu2021swin}, Convolutional vision Transformer (CvT) \cite{wu2021cvt}, Shuffle Transformer \cite{huang2021shuffle}, CrossFormer \cite{wang2021crossformer}, RegionViT \cite{chen2021regionvit} and Focal Transformer models \cite{yang2021focal}. Some of them are hybrid designs (with both convolution and self-attention operations, see Sec.~\ref{Hybrid ViTs with Convolutions}), while others only employ pure self-attention based design (discussed next).

Pyramid ViT (PVT) \cite{wang2021pyramid} is the first hierarchical design for ViT, and proposes a progressive shrinking pyramid and spatial-reduction attention. PVTv2 \cite{wang2021pvtv2} and SegFormer \cite{xie2021segformer} improve original PVT \cite{wang2021pyramid} by introducing overlapping patch embedding, depth-wise convolution, and efficient attention. Swin Transformer \cite{liu2021swin} has a multi-stage hierarchical architecture which computes attention within a local window, by partitioning the window into multiple sub-patches. To capture interactions between different windows (image locations), window partitioning is gradually shifted, along the hierarchy of the network, to capture overlapping regions. 
Focal Transformer models \cite{yang2021focal} is another hierarchical design, where focal self-attention is introduced to simultaneously capture global and local relationships.
Similarly, CrossFormer \cite{wang2021crossformer}  has a hierarchical pyramid structure, and introduces cross-scale embedding module, along-with long short distance attention and dynamic position bias to faithfully capture both local and global visual cues.  RegionViT \cite{chen2021regionvit} proposes a regional-to-local attention to encode hierarchical features. Multi-Scale Vision Longformer \cite{zhang2021multi} also considers a local context in self-attention, but employs the efficient Longformer \cite{beltagy2020longformer} design for self-attention. CrossViT \cite{chen2021crossvit} encodes multi-scale features with two branches (each with multiple transformer blocks), by separately processesing smaller and larger image patches. The information from these two multi-scale bracnches is then fused together using a cross-attention module. 


\subsubsection{Hybrid ViTs with Convolutions}
\label{Hybrid ViTs with Convolutions}
Convolutions do an excellent job at capturing low-level local features in images, and have been explored in multiple hybrid ViT designs, specially at the beginning to ``patchify and tokenize" an input image. For example, Convolutional vision Transformer (CvT) \cite{wu2021cvt} incorporate convolution based projection to capture the spatial structure and low-level details, for tokenization of image patches. CvT has a hierarchical design, where number of tokens is progressively reduced while the token-width is increased, thus imitating the impact of spatial downsampling as in CNNs. Convolution enhanced image Transformers \cite{yuan2021incorporating} employ convolutions based image-to-token module to extract low-level features. Compact Convolutional Transformer (CCT) \cite{hassani2021escaping} introduces a new sequence pooling scheme, and incorporates convolutional blocks (conv-pool-reshape) for tokenization. CCT can be trained from scratch on smaller datasets, \eg, CIFAR10 with $\sim95\%$ accuracy, which is a remarkable property not possible with the traditional ViTs. 

LocalViT \cite{li2021localvit} introduces depthwise convolutions to enhance local features modeling capability of ViTs. LeViT \cite{graham2021levit} (name inspired from LeNet \cite{lecun1989backpropagation}) applies a four-layered CNN block (with $3\times3$ convolutions) at the beginning with progressively increasing channels (3,32,64,128,256). For a $3\times224\times224$ input image, the resulting $256\times14\times14$ output from the CNN block becomes input to a hierarchical ViT. By virtue of its design, LeViT is $5\times$ faster than EfficientNet \cite{tan2019efficientnet} on CPU, at inference. ResT \cite{zhang2021rest} is another hierarchical architecture which applies a CNN block at the beginning for patch-embedding. It incorporates depth-wise convolutions and adaptive position encoding to tackle varying image sizes. A recent approach NesT \cite{zhang2021aggregating} proposes a simple technique to introduce hierarchy in ViTs. NesT divides an image into non-overlapping blocks (each block is further split into patches). It first separately applies local self-attention on patches within each block, and then enables global interaction between blocks by aggregating them into an image space and applying convolution operation, followed by downsampling. The number of blocks is gradually reduced along the hierarchy of the model, while number of local-patches is kept fixed. This simple scheme performs favorably compared with more sophisticated designs \cite{wang2021pvtv2,liu2021swin}, and enables training NesT on smaller datasets (e.g., CIFAR-10) from scratch.

Depthwise Convolution and self-Attention Networks (CoAtNets) \cite{dai2021coatnet} introduce a relative attention module (which combines depthwise convolutions and self-attention), and vertically stack convolution and attention layers. CoAtNets demonstrate an impressive $86\%$ ImageNet top-1 accuracy without extra data (i.e. trained only on ImageNet-1k). Shuffle Transformer \cite{huang2021shuffle} performs self-attention within a window and has depth-wise convolutions between the window-based multi-head self-attention and MLP. It introduces a shuffle operation to build stronger cross-patch connections. 
Co-scale conv-attentional image Transformers (CoaT) \cite{xu2021coscale}, is a hybrid hierarchical pyramid design, with serial and parallel blocks, where the serial block is similar to standard transformer block except for the attention layer replaced with depthwise convolution. The parallel blocks is applied on the output of serial blocks and encodes relationships between tokens at multiple scales using cross-attention. 
Twins \cite{chu2021twins} builds upon PVT \cite{wang2021pyramid} (an attention only pyramid design), by replacing the absolute position embedding in PVT with relative conditional position embedding \cite{chu2021conditional}, and incorporating the separable depth-wise convolutions instead of the standard spatial attention, to capture local and global context of the image. In this sense, the hybrid designs tend to combine the strengths of both convolution and transformer models. TransCNN \cite{liu2021transformer} propose a hierarchical multi-head self attention block, which first learns interactions within small grids (tokens) using self-attention, and then gradually merges the smaller grids into larger grids. The proposed block can then be plugged into existing CNN architectures.

\subsubsection{Self-Supervised Vision Transformers}
 Contrastive learning based self-supervised approaches, which have gained significant success for CNN based vision tasks, have also been investigated for ViTs. Chen \etal \cite{chen2021empirical} evaluate different self-supervised frameworks and propose practical strategies including MoCo v3 (extended from v1/v2 \cite{chen2020improved,he2020momentum}) for stabilized training of self-supervised ViTs. Xie \etal \cite{xie2021self} combine MoCo v2 \cite{he2020momentum} and BYOL \cite{grill2020bootstrap} to train DeiT \cite{touvron2020deit} and SwinTransformer \cite{liu2021swin}. They demonstrate generalization of self-supervised SwinTransformer for dense prediction tasks of detection and segmentation. Self distillation with no labels (DINO) \cite{caron2021emerging} demonstrate that self-supervised ViTs can automatically segment the background pixels of an image, even though they were never trained using pixel-level supervision, a phenomena otherwise not observed in CNNs or fully supervised ViTs. Efficient self-supervised vision transformer (EsViT) \cite{li2021efficient} propose a multi-stage design, where neighboring tokens are gradually merged along the hierarchy of the network, and use DINO for self-supervision. Apart from standard image-level self-supervision as in DINO, they incorporate additional patch-level self-supervision in which correspondence is promoted between similar patches within augmented versions of an image. EsViT demonstrates excellent performance under self-supervision settings, and its off-the-shelf features transfer better than supervised SwinTransformer on 17 out of 18 evaluated datasets.

\subsection{Transformers for Object Detection}
\mh{Transformers based modules have been used for object detection in the following manner: (a) Transformer backbones for feature extraction, with a R-CNN based head for detection (see Sec.~\ref{Hierarchical Multi-Stage ViTs for Dense Prediction}), (b) CNN backbone for visual features and a Transformer based decoder for object detection \cite{carion2020end, zhu2020deformable, wang2021anchor,chen2021pix2seq} (see Sec.~\ref{Detection Transformers with CNN Backbone}, and (c) a purely transformer based design for end-to-end object detection \cite{fang2021look} (see Sec.~\ref{Detection with Pure Transformers}). }


\subsubsection{Detection Transformers with CNN Backbone}
\label{Detection Transformers with CNN Backbone}
Detection Transformer (DETR) \cite{carion2020end} treats object detection as a set prediction task i.e., given a set of image features, the objective is to predict the set of object bounding boxes. The Transformer model enables the prediction of a set of objects (in a single shot) and also allows modeling their relationships. DETR adapts a set loss function which allows bipartite matching between predictions and ground-truth boxes. The main advantage of DETR is that it removes the dependence on hand-crafted modules and operations, such as the RPN (region proposal network) and NMS (non-maximal suppression) commonly used in object detection \cite{ren2016faster,girshick2015fast,he2017mask,redmon2016you,liu2016ssd}. In this manner, the dependence on prior knowledge and careful engineering design is relaxed for complex structured tasks like object detection. 


\begin{figure}[]
    \centering
    \includegraphics[width=1\columnwidth]{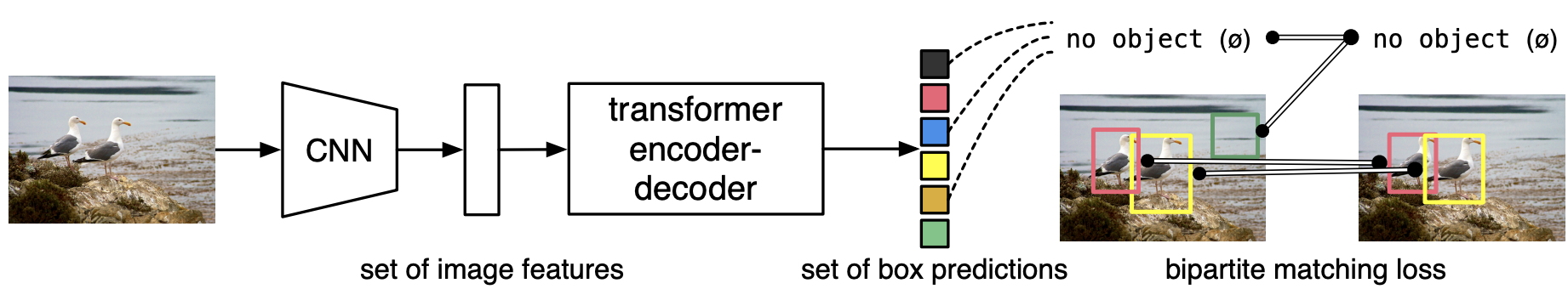}
    \vspace{-0.3cm}
    \caption{\small Detection Transformer (DETR) \cite{carion2020end} treats the object detection task as a set prediction problem and uses the Transformer network to encode relationships between set elements. A bipartite set loss is used to uniquely match the box predictions with the ground-truth boxes (shown on the \emph{right} two columns). In case of no match, a '\emph{no object}' class prediction is selected. Its simple design with minimal problem-specific modifications can beat a carefully built and popular Faster R-CNN model. Figure from \cite{carion2020end}.}
    \label{fig:detr}
\end{figure}

Given spatial feature maps from the CNN backbone, the encoder first flattens the spatial dimensions (see Fig.~\ref{fig:detr}). This gives a sequence of features $d\times n$, where $d$ is the feature dimension and $n = h\times w$ with $h, w$ being the height and width of the spatial feature maps. These features are then encoded and decoded using multi-head self-attention modules as in \cite{vaswani2017attention}. The main difference in the decoding stage is that all boxes are predicted in parallel while \cite{vaswani2017attention} uses an RNN to predict sequence elements one by one. Since the encoder and decoder are permutation invariant, learned positional encodings are used as  the object queries by the decoder to generate different boxes. Note that the spatial structure in a CNN detector (e.g., Faster R-CNN) automatically encodes the positional information. DETR obtains performance comparable to the popular Faster R-CNN model \cite{ren2016faster} which is an impressive feat given its simple design. The DETR has also been extended to interesting applications in other domains, e.g., Cell-DETR \cite{prangemeier2020attention} extends it for instance segmentation of biological cells. A dedicated attention branch is added to obtain instance-wise segmentations in addition box predictions that are enhanced with a CNN decoder to generate accurate instance masks.

The DETR \cite{carion2020end} model successfully combines convolutional networks with Transformers \cite{vaswani2017attention} to remove hand-crafted design requirements and achieves an end-to-end trainable object detection pipeline. However, it struggles to detect small objects and suffers from slow convergence and a relatively high computational cost \cite{zhu2020deformable}. DETR maps images to features space before using the Transformer for the relation modeling. Thus, the computational cost of self-attention grows quadratically with the spatial size of the feature map i.e.,  $\mathcal{O}(H^2W^2C)$, where $H$ and $W$ represent the height and width of the feature map. This inherently puts a limitation on the use of multi-scale hierarchical features \cite{lin2017feature} in DETR training framework which is ultimately important to detect small objects. Furthermore, at the beginning of training, the attention module simply projects uniform attention to all the locations of the feature map and requires a large number of training epochs to tune attention weights to converge to meaningfully sparse locations. This approach contributes to a slow convergence rate of DETR. To mitigate the above-mentioned issues, \cite{zhu2020deformable} proposed a deformable attention module to process the feature maps. Inspired from deformable convolutions \cite{dai2017deformable}, deformable attention module \cite{zhu2020deformable} only attends to sparse set of elements from the whole feature map regardless of its spatial size. This further allows cross-scale aggregation of feature maps with the help of multi-scale attention modules without increasing the computational cost significantly. Deformable DETR not only performs better but its training time also remains 10$\times$ lower than the original DETR model \cite{zhu2020deformable}. 
\mh{Anchor DETR \cite{wang2021anchor} replaces the learnable query tokens in \cite{carion2020end} with anchor-point based queries, such that each query focuses on predicting the object near the anchor point. The anchor points can be fixed on 2D grid, or learned from uniformly distributed points.  Anchor DETR \cite{wang2021anchor} requires 10 $\times$ fewer training epochs with comparable performance.}
\mh{Pix2Seq \cite{chen2021pix2seq} is a generic Transformer-based framework, without any specialized task-specific modules, and learns to directly produce a sequence of tokens with object descriptions (bounding-boxes and class-labels). A quantization and serialization scheme first converts bounding boxes and class-labels into a sequence of discrete tokens. A generic Transformer based encoder-decoder network is then used to generate these tokens in an auto-regressive manner conditioned on previous predictions and image features.}

\subsubsection{Detection with Pure Transformers}\label{Detection with Pure Transformers}
\mh{You Only Look at One Sequence (YOLOS) \cite{fang2021look} is a simple, attention-only architecture directly built upon the ViT \cite{dosovitskiy2020image,vaswani2017attention}. It replaces the class-token in ViT with multiple learnable object query tokens, and the bipartite matching loss is used for object detection similar to \cite{carion2020end}. YOLOS demonstrates the flexibility of ViTs to object detection, in a pure sequence-to-sequence learning manner, with minimal image related 2D inductive biases. In similar spirit, PVT~\cite{wang2021pyramid} is combined with DETR~\cite{carion2020end} to perform object detection with an end-to-end transformer pipeline. We note that it is feasible to combine other recent ViTs with transformer based detection heads as well to create pure ViT based designs \cite{fang2021look}, and we hope to see more such efforts in future.  }

\begin{figure}[]
    \centering
    \includegraphics[width=1\columnwidth]{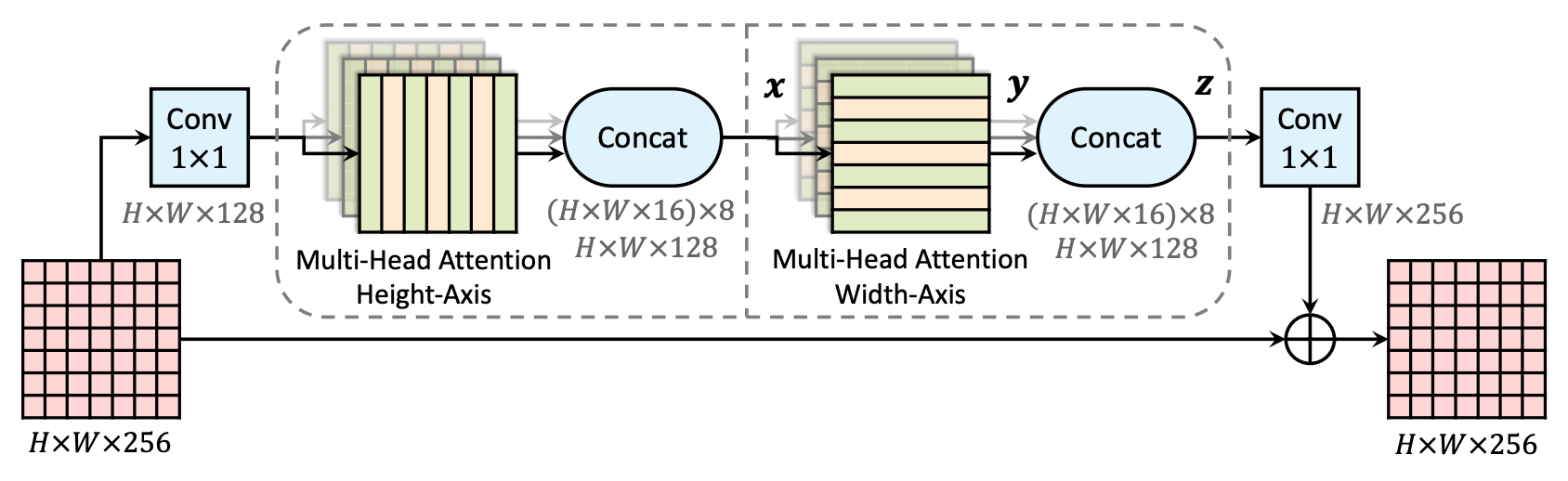}
    \caption{\small Axial attention module \cite{wang2020axial} that sequentially applies multi-head axial attention operations along height and width axes. Image  from \cite{wang2020axial}.}
    \label{fig:axial_attention}
\end{figure}

\subsection{Transformers for Segmentation}
 Self-attention can be leveraged for  dense prediction tasks like image segmentation that requires modeling rich interactions between pixels. Below, we discuss axial self-attention  \cite{wang2020axial}, a cross-modal approach \cite{ye2019cross} that can segment regions corresponding to a given language expression, and ViTs based segmentation architectures \cite{zheng2021rethinking,xie2021segformer,strudel2021segmenter}.

Panoptic segmentation~\cite{kirillov2019panoptic} aims to jointly solve the otherwise distinct tasks of semantic segmentation and instance segmentation by assigning each pixel a semantic label and an instance id. Global context can provide useful cues to deal with such a complex visual understanding task. Self-attention is effective at modeling long-range contextual information, albeit applying it to large inputs for a dense prediction task like panoptic segmentation is prohibitively expensive.  A naive solution is to apply self-attention either to downsampled inputs or to limited regions around each pixel~\cite{parmar2019stand}. Even after introducing these constraints, the self-attention still has quadratic complexity and sacrifices the global context.
To tackle these issues, Wang \etal~\cite{wang2020axial} propose the position-sensitive axial-attention where the 2D self-attention mechanism is reformulated as two 1D axial-attention layers, applied to height-axis and width-axis sequentially (see Fig.~\ref{fig:axial_attention}). The axial-attention is compute efficient and enables models to capture the full-image context. It achieves competitive performance for the panoptic segmentation task on COCO~\cite{lin2014coco}, Mapillary Vistas~\cite{neuhold2017mapillary}, and Cityscapes~\cite{cordts2016cityscapes} benchmarks and for the image classification on ImageNet dataset~\cite{deng2009imagenet}.

Cross-modal Self-attention (CMSA) \cite{ye2019cross} encodes long-range multi-modal dependencies between linguistic and visual features for \textit{referring image segmentation task}, that aims to segment entities in an image referred by a language description.
For this purpose, a set of cross-modal features is obtained by concatenating image features with each word embedding and the spatial coordinate features. The self-attention operates on these features and generates attention over the image corresponding to each word in the sentence. The segmentation network then performs self-attention at multiple spatial levels and uses a gated multi-level fusion module to refine segmentation masks via information exchange across multi-resolution features. A binary CE loss is used to train the overall model that achieves good improvements on UNC \cite{yu2016modeling}, G-Ref \cite{mao2016generation} and ReferIt \cite{kazemzadeh2014referitgame} datasets.


\mh{
While the segmentation approaches discussed above insert self-attention in their CNN based architectures, some recent works have proposed transformer based encoder-decoder architectures. Segmentation Transformer (SETR) \cite{zheng2021rethinking} has a ViT encoder, and two decoder designs based upon progressive upsampling, and multi-level feature aggregation.
SegFormer \cite{xie2021segformer} has a hierarchical pyramid ViT \cite{wang2021pyramid} (without position encoding) as an encoder, and a simple MLP based decoder with upsampling operation to get the segmentation mask. 
Segmenter \cite{strudel2021segmenter} uses ViT encoder to extract image features, and the decoder is a mask Transformer module which predicts segmentation masks, using learnable mask tokens and image-patch tokens as inputs. The authors also propose a baseline linear decoder which projects the patch-embeddings to classification space, thus producing coarse patch-level labels.
}

\subsection{Transformers for Image and Scene Generation}

Here, we discuss Transformer-based architectures \cite{parmar2018imagetransformer,chen2020generative,esser2020taming, jiang2021transgan, bhunia2021handwriting, wang2020sceneformer} for image synthesis, which is interesting from the perspective of generative modeling and learning unsupervised representations for down-stream tasks.

\begin{figure}[]
     \centering
     \includegraphics[clip=true, trim=0mm 0mm 1mm 2.0mm, width=1\columnwidth]{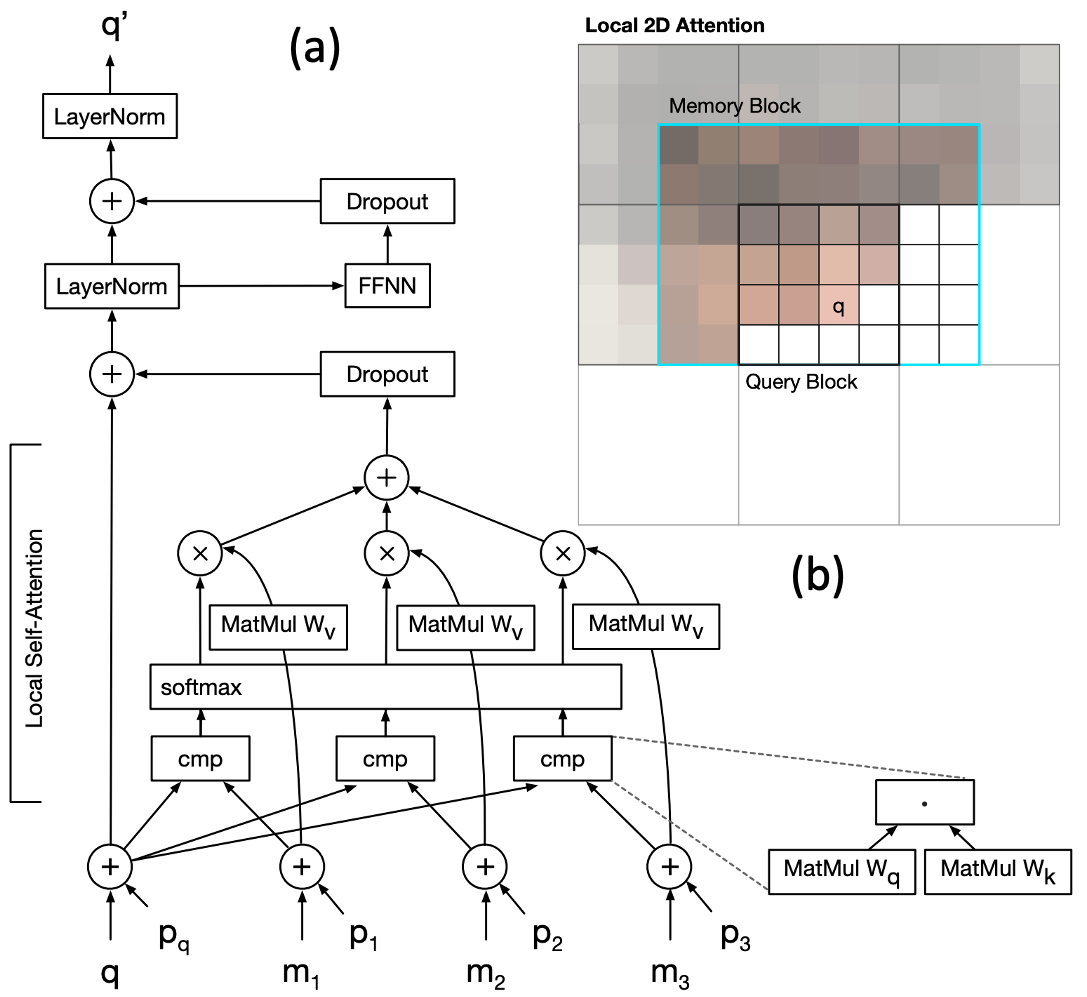}
     \caption{\small (a) Self-attention block in Image Transformer \cite{parmar2018imagetransformer}. Given one channel for a pixel $q$, the block attends to the memory of previous synthesized pixels ($m_i$), followed by a feed-forward sub-network. Positional encodings $p_i$ are added in the first layer. (b) The operation performed in Local Self-Attention (example of a 2D case is shown). The image is partitioned into a grid of spatial blocks known as query blocks. In the self-attention operation, each pixel in a query block attends to all pixels in the memory block (shown in cyan rectangle). White grid locations show masked inputs that have zero-contribution towards the self-attention.}
     \label{fig:img_transformer}
 \end{figure}

Parmar \etal \cite{parmar2018imagetransformer} develop an image generation model that can sequentially predict each pixel of an output image given its previously generated pixels (Fig.~\ref{fig:img_transformer}). Their approach models the joint distribution of the image pixels by factorizing it as a product of pixel-wise conditional distributions. Previously developed auto-regressive models for this task, such as the PixelCNN~\cite{van2016conditional}, suffer from a limited receptive field which hinders in modeling long term relationships in an image \eg, part relationships or occlusions. Using self-attention,  \cite{parmar2018imagetransformer} enhances the receptive field without incurring a high computational cost (\eg, effective receptive field up to 256 pixels can be achieved as compared to 25 pixels of PixelCNN~\cite{van2016conditional}). The generative pipeline was also tested on conditional generation tasks \eg, image super-resolution, image completion, and denoising.

 
Inspired by the success of GPT model \cite{radford2019language} in the language domain, image GPT (iGPT) \cite{chen2020generative} demonstrated that such models can be directly used for image generation tasks, and to learn strong features for downstream vision tasks (e.g., image classification).  Specifically, iGPT trains GPT v2 model \cite{radford2019language} on flattened image sequences (1D pixel arrays) and shows that it can generate plausible image outputs without any external supervision. The generated samples depict the model's ability to understand spatial relationships between pixels and high-level attributes such as object classes, texture, and scale. Notably, the design does not use any image-specific knowledge in the design (e.g., the 2D position embeddings used in Image Transformer \cite{parmar2018imagetransformer}).
The features learned with iGPT's unsupervised training mechanism compete impressively against other unsupervised approaches, achieving state-of-the-art performance on CIFAR-10/100 \cite{krizhevsky2009learning} and STL \cite{coates2011analysis} datasets while performing comparably to SimCLR (a contrastive learning approach) \cite{chen2020simple} on ImageNet dataset. This is an astounding result, since the iGPT architecture is exactly the same as used for language modeling tasks, and therefore it does not incorporate any prior domain-specific knowledge. Notably, the competing unsupervised CNN based solutions widely adopt such priors in the form of architectural design, attention mechanisms, loss functions, and regularization \cite{bachman2019learning, henaff2019data,he2020momentum,tian2019contrastive,khan2018guide}. However, on the downside, iGPT has a high compute cost \eg, iGPT-L version has roughly $36\times$ high training cost compared to MoCo \cite{he2020momentum} which is a state of the art self-supervised feature learning approach. For this reason, the training was generally limited to low-resolution of $\leq 64\times 64$, while convolutional architectures can effectively learn from high-resolution inputs.

Transformers typically incur a high compute cost when applied on high-dimensional sequences. To overcome this limitation, Esser \etal \cite{esser2020taming} proposed to include inductive biases (commonly used in the CNNs) alongside Transformers to improve their efficiency. Specifically, local connectivity and spatial invariance biases inbuilt in the CNN structure are leveraged by learning a rich dictionary of visual patterns (using a Generative Adversarial approach). A Transformer is then used to learn the long-range interactions between the dictionary items to generate the outputs. In turn, they develop a conditional image generation model capable of producing very high-resolution images (up to megapixel range) using Transformers. This is the first work that demonstrates the application of Transformers to generate such high-resolution images.

Generative Adversarial Networks (GANs) \cite{Goodfellow2014} with CNNs as default backbone have been very successful for visually appealing image synthesis \cite{radford2015unsupervised,gao2020adversarialnas,karras2020analyzing}. TransGAN \cite{jiang2021transgan} builds a strong GAN model, free of any convolution operation, with both generator and discriminator based upon the Transformer model \cite{vaswani2017attention}. 
The architecture of both generator and discriminator is based upon the encoder in original Transformer model \cite{vaswani2017attention}. For memory efficiency, the generator contains multiple stages, with up-sampling modules in-between, which gradually increase the resolution of feature maps (input sequence length) while reducing the embedding dimension. The discriminator of TransGAN takes flattened image-patches as tokens similar to \cite{dosovitskiy2020image}.
Authors introduce different training techniques including data augmentation, training with an auxiliary task and injecting locality to self-attention to scale-up their model for high quality image synthesis \cite{esser2020taming}. The TransGAN model achieves state-of-the-art results in terms of Inception Score and Fr\'echet Inception Distance (FID) on STL-10 and performs favorably compared with their CNN-based GAN counterparts on other datasets.


 Unlike previous image generation methods \cite{parmar2018imagetransformer,chen2020generative,esser2020taming}, which directly predict image outputs,  \cite{wang2020sceneformer} learns to generate parameters of 3D objects to be placed in a given scene. Specifically, SceneFormer \cite{wang2020sceneformer} studies the 3D room layout conditioned scene generation task. Given the empty room shape, \cite{wang2020sceneformer}  can propose new object configurations in the room while maintaining realism. Remarkably, the model does not use any appearance information and only learns to generate new scenes by modeling the inter-object relationships using self-attention in Transformers. Similar to how a Transformer operates on a sentence, it is applied to a sequence of objects to predict the next suitable object in a scene. Specifically, the size, pose, location, and category of the next object is predicted by the Transformer model. A start token indicates the initiation of inference and the number of output token indicate the objects generated by the model in a sequence. The authors also explore generating new scenes given a textual description of the room layout. The independence from the appearance makes the approach efficient, enabling interactive scene generation.

\begin{figure*}
\centering
    \begin{subfigure}[t]{0.13\textwidth}
      \includegraphics[width=\textwidth]{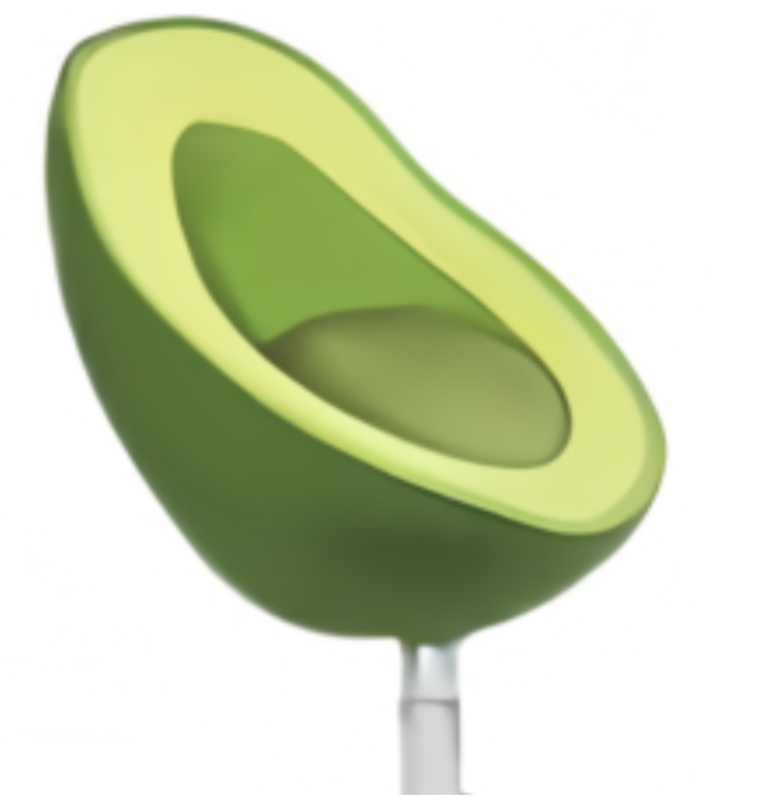}
      \caption{\small}
      \label{fig:dalle scratch}
    \end{subfigure}
    \begin{subfigure}[t]{0.13\textwidth}
      \includegraphics[width=\textwidth]{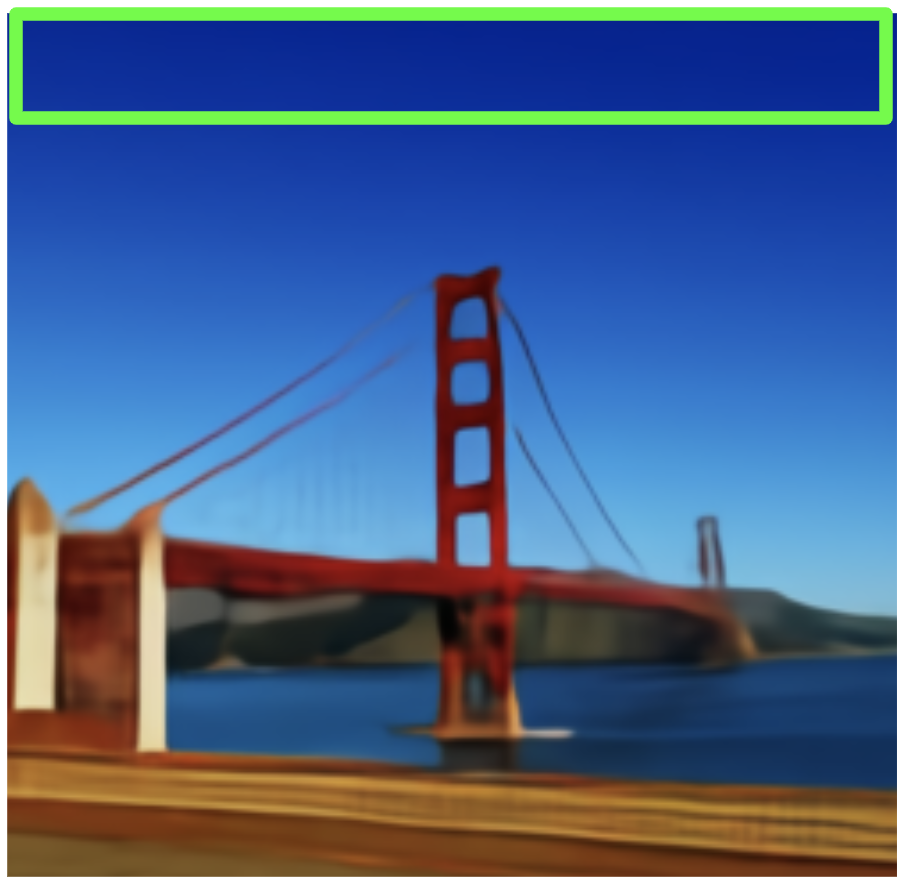}
      \caption{\small }
      \label{fig:dalle img_comp}
    \end{subfigure}
    \begin{subfigure}[t]{0.13\textwidth}
      \includegraphics[width=\textwidth]{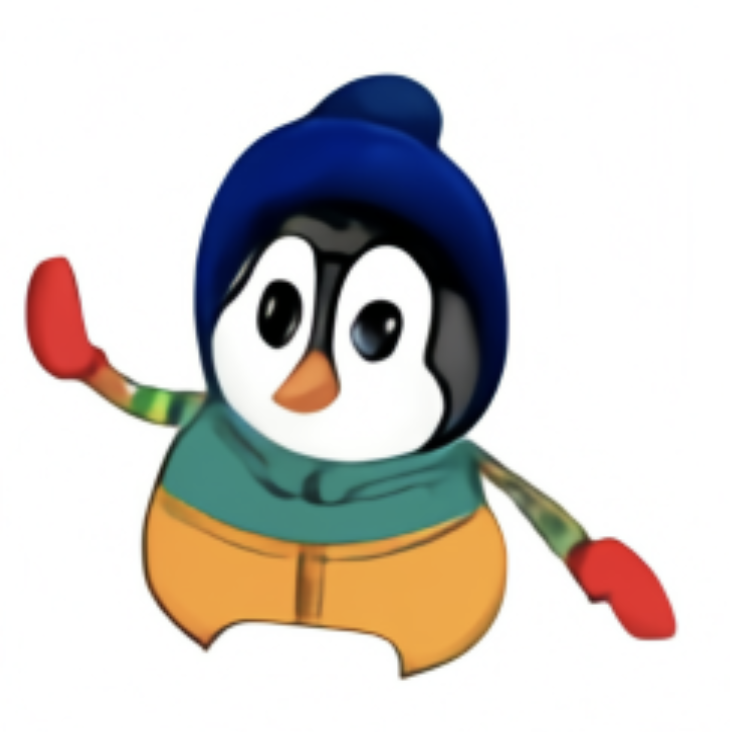}
      \caption{\small }
      \label{fig:dalle multi-attr}
    \end{subfigure}
    \begin{subfigure}[t]{0.13\textwidth}
      \includegraphics[width=\textwidth]{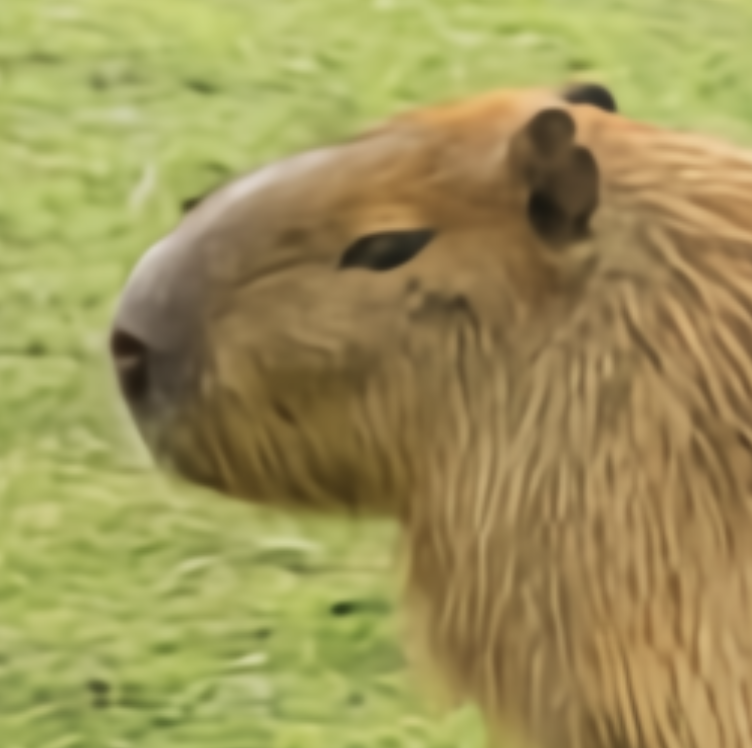}
      \caption{\small }
      \label{fig:dalle viewpoint}
    \end{subfigure}
    \begin{subfigure}[t]{0.13\textwidth}
      \includegraphics[width=\textwidth]{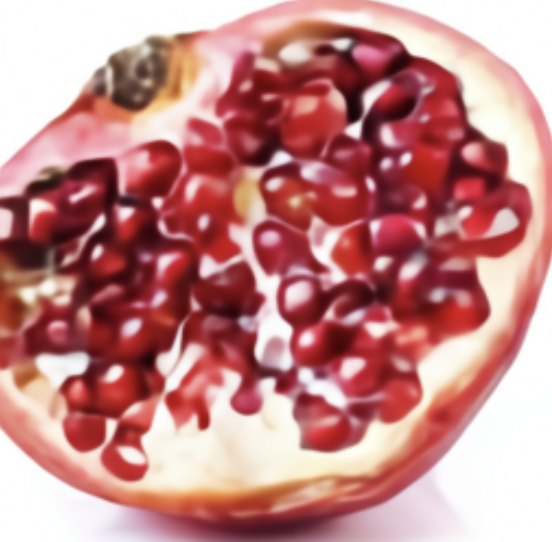}
      \caption{\small}
      \label{fig:dalle cross-section}
    \end{subfigure}
    \begin{subfigure}[t]{0.13\textwidth}
      \includegraphics[width=\textwidth]{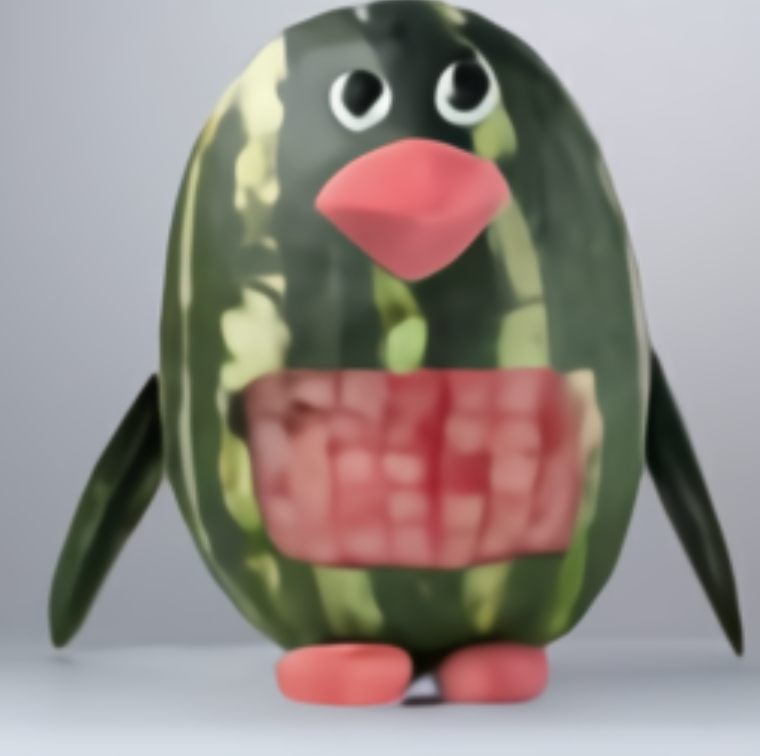}
      \caption{\small}
      \label{fig:dalle unrelated concepts}
    \end{subfigure}
    \begin{subfigure}[t]{0.13\textwidth}
      \includegraphics[width=\textwidth]{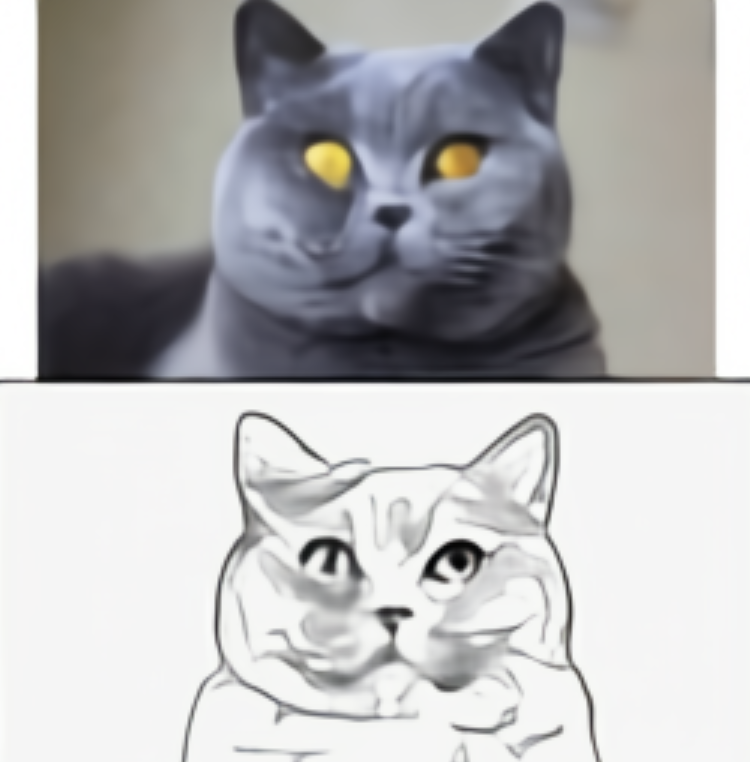}
      \caption{\small}
      \label{fig:dalle img2img}
    \end{subfigure}
\vspace{-0.3cm}
\caption{\small Images generated by DALL·E~\cite{Ramesh2021dalle} from the following text prompts. (a) \emph{An armchair in the shape of an avocado.} (b) \emph{A photo of San Francisco's golden gate bridge.} Given a part of the image (in green box), DALL·E performs the image completion. (c) \emph{An emoji of a baby penguin wearing a blue hat, red gloves, green shirt, and yellow pants.} (d) \emph{An extreme close-up view of a capybara sitting in a field.} (e) \textit{ A cross-section view of a pomegranate.} (f) \emph{A penguin made of watermelon.} (g) \emph{The exact same cat on the top as a sketch on the bottom.} }
\end{figure*}

The task of generating realistic images from text is interesting and practically valuable (\eg, for artistic content creation), but at the same time highly challenging. Prior text-to-image synthesis approaches~\cite{reed2016generative,zhang2017stackgan,zhang2018stackgan++,xu2018attngan} are mostly based on GANs~\cite{Goodfellow2014}. Although these methods produce encouraging results, they are far from being photo-realistic. 
Ramesh~\etal~\cite{Ramesh2021dalle} recently proposed DALL·E which is a Transformer model capable of generating high-fidelity images from a given text description. 
DALL·E model has 12 billion parameters and it is trained on a large set of text-image pairs taken from the internet. 
Before training, images are first resized to 256$\times$256 resolution, and subsequently compressed to a 32$\times$32 grid of latent codes using a pre-trained discrete variational autoencoder~\cite{kingma2013auto,razavi2019generating}. 
DALL·E takes as input a single stream of 1280 tokens (256 for the text and 1024 for the image), and is trained to generate all other tokens autoregressively (one after another). It provides flexibility to generate images either from scratch (Fig.~\ref{fig:dalle scratch}) or by extending existing images (Fig.~\ref{fig:dalle img_comp}), while staying faithful to the text caption.

The authors demonstrate the effectiveness of DALL·E by creating images from text describing a wide variety of real and fictional concepts. While generating images purely from textural captions, DALL·E shows impressive performance at controlling multiple objects and their attributes (Fig.~\ref{fig:dalle multi-attr}), rendering certain viewpoint (Fig.~\ref{fig:dalle viewpoint}), capturing object's internal structure (Fig.~\ref{fig:dalle cross-section}), and combining unrelated objects (Fig.~\ref{fig:dalle unrelated concepts}). 
Furthermore, DALL·E can perform image-to-image translation (Fig.~\ref{fig:dalle img2img}) guided by the input text.

\subsection{Transformers for Low-level Vision}
After witnessing the success of Transformer models in high-level vision problems, numerous Transformer-based methods have been proposed for low-level vision tasks, including image super-resolution~\cite{yang2020superresolution,liang2021swinir,chen2020ipt}, denoising~\cite{wang2021uformer,chen2020ipt}, deraining~\cite{wang2021uformer,chen2020ipt}, and colorization~\cite{anonymous2021colorization}.  
Image restoration requires pixel-to-pixel correspondence from the input to the output images. One major goal of restoration algorithms is to preserve desired fine image details (such as edges and texture) in the restored images. CNNs achieve this by employing a single-scale architecture design that does not involve any downsampling operation. Since the computational complexity of self-attention in Transformer models increases quadratically with number of image patches, it is infeasible to develop Transformer model that can operate on single-scale feature processing pipeline. Consequently, these Transformer-based image restoration models make use of various strategies to reduce the computational burden, such as computing attention on local image windows~\cite{liang2021swinir}, performing spatial reduction attention~\cite{lu2021efficientSR}, and employing encoder-decoder design~\cite{chen2020ipt, wang2021uformer}. Here, we briefly discuss a few image restoration Transformer models.


\subsubsection{Transformers for Image Processing Tasks}
Top performing algorithms for high-level computer vision tasks such as object detection and semantic segmentation often employ backbone models that are pre-trained on large-scale datasets \eg, ImageNet. In contrast, algorithms for low-level vision tasks such as image denoising, super-resolution, and deraining are directly trained on task-specific data, thereby suffer from these limitations: \textbf{(i)} small number of images available in task-specific datasets (\eg, the commonly used DIV2K dataset for image super-resolution contains only 2000 images), \textbf{(ii)} the model trained for one image processing task does not adapt well to other related tasks.

Chen \etal~\cite{chen2020ipt} propose a pre-trained model based on Transformer architecture, named as Image Processing Transformer (IPT). It is capable of performing various image restoration tasks such as super-resolution, denoising, and deraining. 
The overall architecture of IPT consists of multi-heads and multi-tails to deal with different tasks separately, and a shared encoder-decoder Transformer body. 
Since exploiting Transformers at full potential requires training on large-scale data, \cite{chen2020ipt} takes the clean (ground-truth) images from the ImageNet benchmark and synthesize their degraded versions for different tasks. For example, bicubic interpolation is used for generating low-resolution images, additive white Gaussian noise is added to prepare noisy data, and hand-crafted rain streaks are applied to obtain rainy images. In total, 10 million images are used to pre-train the IPT model. During training, each task-specific head takes as input a degraded image and generates visual features. These feature maps are divided into small crops and subsequently flattened before feeding them to the Transformer encoder (whose architecture is the same as \cite{vaswani2017attention}). The outputs of the encoder along with the task-specific embeddings are given as input to the Transformer decoder. The features from the decoder output are reshaped and passed to the multi-tail that yields restored images. The IPT model is optimized with L$_1$ loss. Experimental results show that the pre-trained IPT model, when fine-tuned for a specific low-level vision task, can provide significant performance gains over the state-of-the-art methods \cite{RCAN,dai2019second,niu2020single}.


\subsubsection{Transformers for Super-Resolution}
Recent years have seen major performance breakthroughs for super-resolution (SR) due to convolutional neural networks (CNNs). Principally, the quality of super-resolved images generated by CNNs is dependent on the choice of optimization objective. While the SR methods \cite{EDSR,tai2017image,han2018image,RCAN,zhang2020residual} that are based on pixel-wise loss functions (\eg, L1, MSE, etc.) yield impressive results in terms of image fidelity metrics such as PSNR and SSIM, they struggle to recover fine texture details and often produce images that are overly-smooth and perceptually less pleasant.
Further, \emph{perceptual} SR approaches \cite{wang2018esrgan,park2018srfeat,sajjadi2017enhancenet,SRResNet,ledig2017photo}, in addition to per-pixel loss, employ adversarial loss \cite{Goodfellow2014} and perceptual loss \cite{Johnson2016} based on deep features extracted from pre-trained CNNs. While these methods generate images that are sharp, visually pleasant, and perceptually plausible, they show a substantial decrease in reconstruction accuracy measured in PSNR/SSIM. Moreover, the perceptual SR algorithms have a tendency to hallucinate fake textures and cause artifacts. The above mentioned SR approaches follow two distinct (but conflicting) research directions: one maximizing the reconstruction accuracy and the other maximizing the perceptual quality, but never both.

\begin{figure}[htp]
    \centering
    \includegraphics[width=0.75\columnwidth]{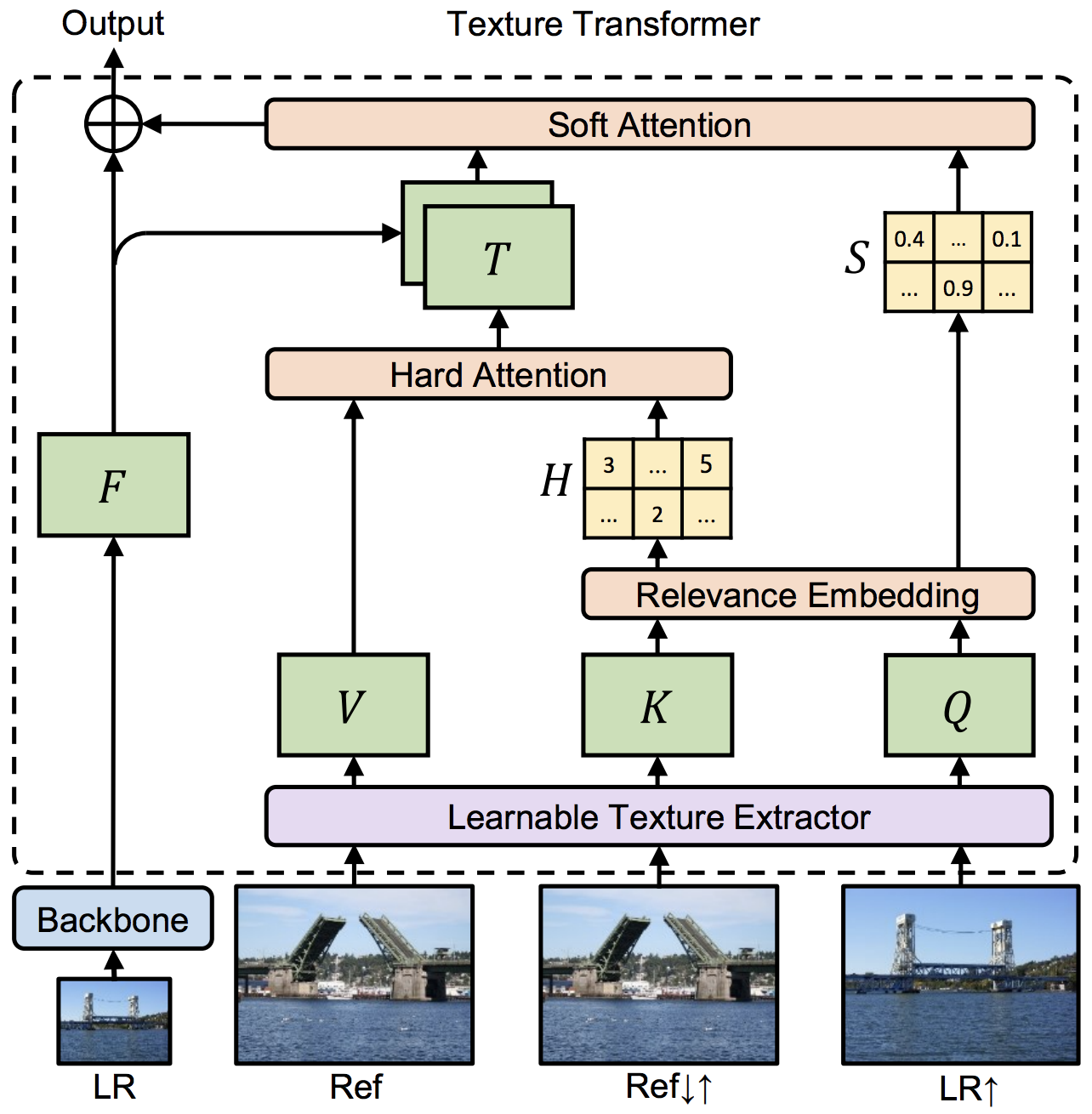}
    \vspace{-0.2cm}
    \caption{\small Diagram of the texture Transformer module. $Q$ (query), $K$ (key) and $V$ (value) represent texture features extracted from a (bicubic upsampled) low-resolution image, a sequentially down/upsampled reference image, and an original reference image, respectively. The relevance embedding aims to estimate similarity between low-resolution and reference images. $H$ and $S$ respectively denote hard and soft attentions computed from relevance embedding. $T$ indicates high-resolution texture features that are then transferred to the features $F$ of low-resolution image. Figure is from \cite{yang2020superresolution}.}
    \label{fig:ttsr}
\end{figure}

To alleviate the trade-off between perceptual reproduction and accurate reproduction, Yang \etal \cite{yang2020superresolution} propose a Transformer network (TTSR) for super-resolution.
During training, TTSR uses paired LR-HR images, as well as reference (Ref) images with similar content as of LR images.
TTSR learns to search relevant regions in the Ref image and transfers rich textures to help super-resolving the input LR image.
The texture Transformer module of TTSR method (see Fig.~\ref{fig:ttsr}) consists of four core components: (1) \emph{Learnable texture extractor:} takes as input LR$\uparrow$, Ref$\downarrow \uparrow$, and Ref images, and generates texture features query (Q), key (K), and value (V), respectively. Here, $\uparrow$ denotes bicubic upsampling operation, and $\downarrow \uparrow$ represents bicubic down-sampling followed by an upsampling operation. (2) \emph{Relevance embedding:} first unfolds Q and K into patches and then computes the similarity of each patch in Q with each patch in K in order to generate hard and soft attention maps. (3) \emph{Hard-attention:} transfers HR texture features from V to (LR features) Q using the hard attention map. (4) \emph{Soft-attention:} further enhances relevant features while suppressing less relevant ones.

{While TTSR~\cite{yang2020superresolution} method deals with reference-based image super-resolution, most of the research is conducted on single image super-resolution problem in which only LR-HR paired images are available. Since the computational complexity of the original self-attention operation is prohibitively high for high-resolution images, recently a few efficient transformer models have been proposed that employ window-based attention (SwinIR~\cite{liang2021swinir}) and spatial resolution reduction operation in attention module (ESRT~\cite{lu2021efficientSR}) to perform super-resolution.}

\subsubsection{Colorization Transformer}
Given a grayscale image, colorization seeks to produce the corresponding colorized sample. It is a one-to-many task as for a given grayscale input, there exist many possibilities in the colorized output space. The challenging nature of this task requires probabilistic models capable of producing multiple colorized output samples. Colorization Transformer \cite{anonymous2021colorization} is a probabilistic model based on conditional attention mechanism \cite{ho2019axial}. It divides the image colorization task into three sub-problems 
and proposes to solve each task sequentially by a different Transformer network. The authors first train a Transformer network to map a low-resolution grey-scale image to a 3-bit low-resolution colored image. Low-resolution images in turn allow training of larger models. The 3-bit low-resolution colored image is then upsampled to an 8-bit RGB sample by another Transformer network in the second stage of training. Finally, a third stage Transformer is trained to increase the spatial resolution of the 8-bit RGB sample produced by the second-stage Transformer. Self-attention used in the colorization Transformer is based on row/column attention layers introduced in \cite{ho2019axial}. These layers capture the interaction between each pixel of an input image while being computationally less costly. The row-wise attention layer applies self-attention to all pixels in a given row, while the column-wise attention layer considers pixels only in a given column of an image.  This work \cite{anonymous2021colorization} is the first successful application of Transformers trained to colorize grey-scale images at high (256$\times$256) resolution.


\begin{figure*}[]
    \centering
    \includegraphics[width=\textwidth]{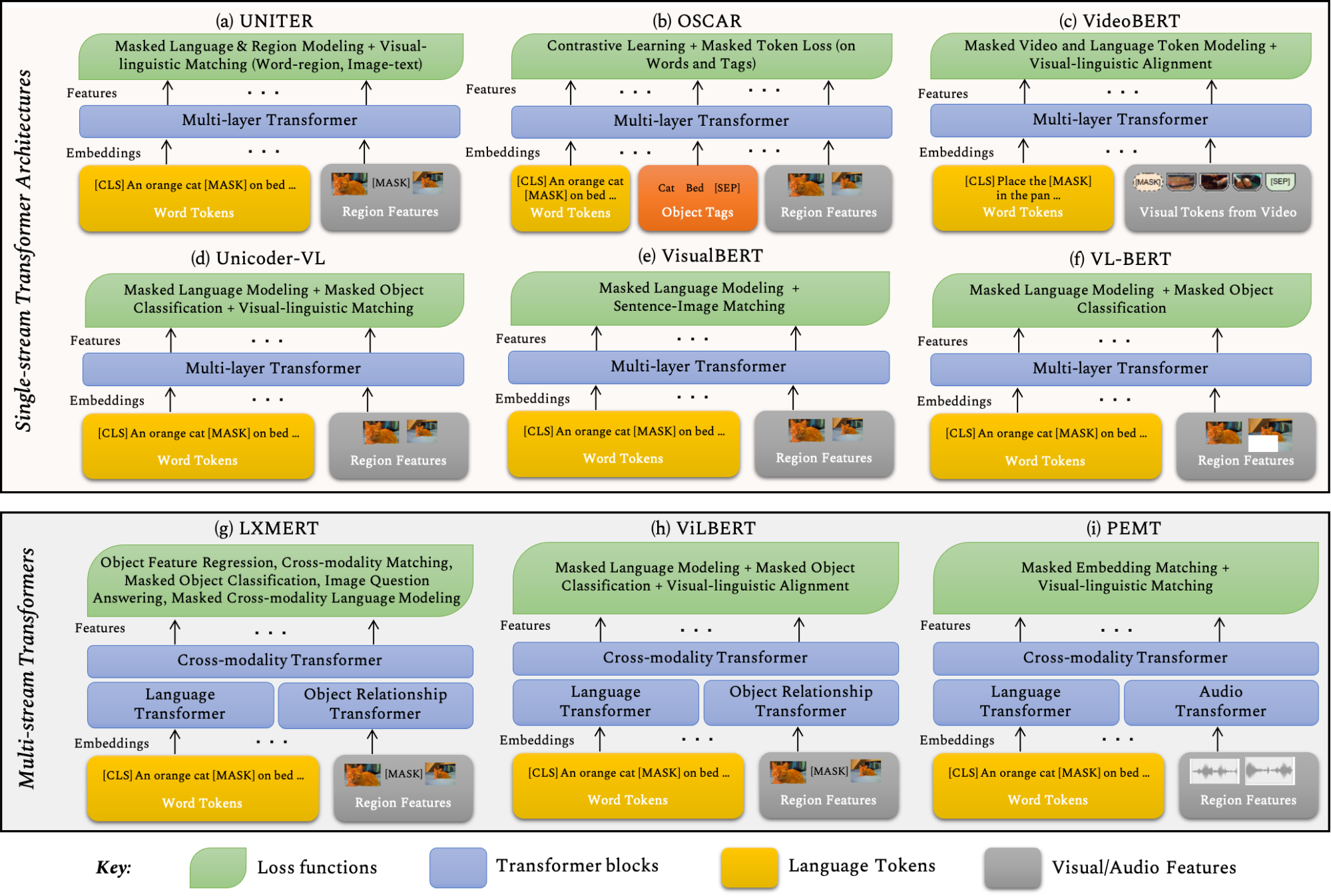}
    \vspace{-0.3cm}
    \caption{\small An overview of Transformer models used for multi-modal tasks in computer vision. The Transformer designs in this category can be grouped into single-stream (UNITER \cite{chen2020uniter}, OSCAR \cite{li2020oscar}, VideoBERT \cite{sun2019videobert}, Unicoder-VL \cite{li2020unicoder}, VisualBERT \cite{li2019visualbert} and VL-BERT \cite{su2019vl}) and dual-stream architectures (LXMERT \cite{tan2019lxmert}, ViLBERT \cite{lu2019vilbert} and PEMT \cite{lee2020parameter}). A key distinction between models is the choice of loss functions. While most of the multi-modal methods are focused on images as visual data, VideoBERT \cite{sun2019videobert} and PEMT \cite{lee2020parameter} are designed to work on video streams and leverage unique modalities e.g., audio signals in videos \cite{lee2020parameter}.}
    \label{fig:multi-modal-archs}
\end{figure*}


\subsection{Transformers for Multi-Modal Tasks}
Transformer models have also been extensively used for vision-language tasks such as visual question answering (VQA) \cite{antol2015vqa}, visual commonsense reasoning (VSR) \cite{zellers2019vcr}, cross-modal retrieval \cite{lee2018stacked} and image captioning \cite{vinyals2015show}. Several works in this direction target effective vision-language pre-training (VLP) on large-scale multi-modal datasets to learn generic representations that effectively encode cross-modality relationships (\eg, grounding semantic attributes of a person in a given image). These representations can then be transferred to downstream tasks, often obtaining state of the art results. \sk{Notably, several of these models still use CNNs as vision backbone to extract visual features while Transformers are used mainly used to encode text followed by the fusion of language and visual features.} Such models generally apply the vanilla multi-layer Transformer \cite{vaswani2017attention} with multi-modal inputs and do not introduce fundamental changes to the core attention block. However, their main distinction is in the configuration of Transformers and the loss functions, based on which we categorize them into: (a) Multi-stream Transformers (see Sec.~\ref{Multi-stream Transformers}) and (b) Single-stream Transformers (see Sec.~\ref{Single-stream Transformers}). The \emph{single-stream} designs feed the \emph{multi-modal} inputs to a single Transformer while the multi-stream designs first use independent Transformers for each modality and later learn cross-modal representations using another Transformer (see Fig.~\ref{fig:multi-modal-archs}). Besides these vision language pretraining methods, we also explain visual grounding approaches towards the end of this section (see Sec.~\ref{Transformers for Visual Grounding}).

\subsubsection{Multi-stream Transformers}\label{Multi-stream Transformers}
Vision and Language BERT (ViLBERT) \cite{li2019visualbert} was the first extension of the BERT model to the multi-modal domain. The goal was to learn representations that can jointly model images and natural language. For this purpose, ViLBERT developed a two-stream architecture where each stream is dedicated to model the vision or language inputs (Fig.~\ref{fig:multi-modal-archs}-h). The architecture of both parallel streams is a series of Transformer blocks similar to the BERT model. Subsequently, co-attentional Transformer layers are applied to learn cross-modal relationships. The co-attentional framework is very simple. Query, key, and value matrices are computed for each modality in the standard way \cite{vaswani2017attention} and then key-value pairs for one modality are passed on to the other modality's attention head. 

ViLBERT applies VLP on a set of proxy tasks defined on the Conceptual Concepts dataset (with 3.3M images with weak captions) and later fine-tune the model on downstream tasks such as VQA. The pre-training phase operates in a self-supervised manner, i.e., pretext tasks are created without manual labeling on the large-scale unlabelled dataset. These pretext tasks include predicting whether the text and image inputs are related and predicting the semantics of masked image regions and textual inputs (\eg, similar to reconstructing masked words in text in the BERT model \cite{devlin2018bert}). This way, the model learns the inherent structure in the data during pre-training and also models cross-domain associations. With evaluations on several tasks, \cite{sun2019videobert} demonstrated that a two-stream model can perform better than a single-stream model that uses shared parameters to model both language and vision domains \cite{sun2019videobert}.

Similar to ViLBERT \cite{lu2019vilbert}, Learning Cross-Modality Encoder Representations from Transformers (LXMERT) \cite{tan2019lxmert} also uses a two-stream architecture based on BERT framework. The main difference lies in the object-relationship encoder that is used to model the visual features instead of simple image-level features used in ViLBERT. The information in two streams is then fused across modalities using cross-attention blocks similar to \cite{lu2019vilbert}.

Compared to two pre-texts tasks used for VLP in \cite{lu2019vilbert}, LXMERT uses five pre-training tasks including masked object and language prediction, cross-modality matching, and visual question answering (Fig.~\ref{fig:multi-modal-archs}-g). The pre-trained model is fine-tuned on the VQA task, however, a high similarity between pre-training and fine-tuned tasks raises questions on the generalizability of the learned representations to new tasks. To this end, the authors conducted generalization experiments on Visual Reasoning for Real (NLVR) task \cite{suhr2018corpus} demonstrating impressive improvements on novel tasks. 

Lee \etal
\cite{lee2020parameter} note that the multi-modal representation learning approaches like VideoBERT \cite{sun2019videobert} and ViLBERT \cite{lu2019vilbert} generally keep the language processing part fixed to a pre-trained model (\eg, BERT \cite{devlin2018bert}) to reduce training complexity.  For the first time in the literature, they propose to learn an end-to-end multi-modal bidirectional Transformer model called PEMT on audio-visual data from unlabeled videos.  First, short-term (\eg, 1-3 seconds) video dynamics are encoded using CNNs, followed by a modality-specific Transformer (audio/visual) to model long-term dependencies (\eg, 30 seconds). A multi-modal Transformer is then applied to the modality-specific Transformer outputs to exchange information across visual-linguistic domains. However, learning such a model in a naive form would incur huge memory requirements. To reduce parametric complexity, the parameters are shared across layers within each Transformer which leads upto 80\% parameter reduction.
The Transformer is trained using a contrastive learning approach based on a content-aware negative sampling (Fig.~\ref{fig:multi-modal-archs}-i).  Specifically, the model uses the features obtained from CNNs learned during the training phase to select negative samples that are visually similar to the positive instances. This work also compares various fusion strategies adopted in earlier works such as early (VideoBERT \cite{sun2019videobert} and VL-BERT \cite{su2019vl}), mid-level (ViL-BERT \cite{lu2019vilbert} and  LXMERT \cite{tan2019lxmert}) and late fusion mechanisms and shows that the mid-level fusion is the optimal choice. The proposed model is pre-trained on Kinetics-700 \cite{carreira2019short} dataset and later fine-tuned on downstream video classification tasks such as short video classification on UCF101 \cite{soomro2012ucf101}, audio classification on ESC50 \cite{gemmeke2017audio} and long-term action recognition on Charades \cite{sigurdsson2016hollywood} and Kinetics-Sounds \cite{arandjelovic2017look} datasets.

\begin{figure}[]
    \centering
    \includegraphics[width=1\columnwidth]{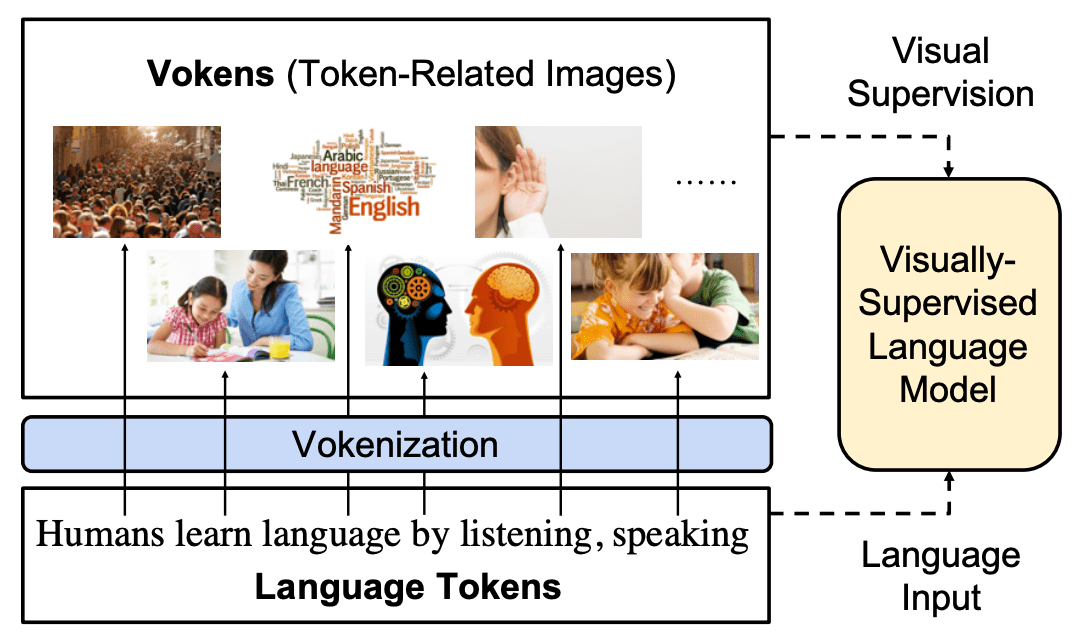}
    \caption{\small Visualized tokens (Vokens) \cite{tan2020vokenization}: A language model is visually supervised using closely related images that leads to better feature representations from the pretrained model. Figure from \cite{tan2020vokenization}. }
    \label{fig:voken}
\end{figure}

Tan and Bansal
\cite{tan2020vokenization} introduce the concept of `\emph{vokens}' (images related to language tokens extracted from sentences). The vokens (visualized tokens) provide visual supervision to the language model to learn better features. The motivation is that humans learn languages by correlating visual information with semantic concepts. In a similar spirit to other self-supervised language representation learning methods \cite{lu2019vilbert,devlin2018bert}, they learn representations by defining an auxiliary task of voken-prediction task.
Since the existing datasets encode limited visually grounded tokens, they propose a vokenization method to map language tokens to visual vokens, as illustrated in Fig.~\ref{fig:voken}. 
The approach uses language-based retrieval for such a mapping and transfers a model trained on a small labeled dataset (MS-COCO) to a large dataset (Wikipedia). Furthermore, it was ensured that the sentence-wide context is considered to obtain the token-voken mapping. The resulting model trained using generated tokens outperforms the state of the art BERT model on a diverse set of NLP tasks. In this sense, the proposed model does not evaluate vision tasks, however, uses vision as a useful grounding cue to train the language model, hence we include it in the multi-modal representation learning group.

Vision-and-Language Navigation (VLN) aims to predict a navigation plan on a map based on the vision and language inputs. Transformer models were used earlier in 
\cite{hao2020towards,majumdar2020improving} for VLN task. These works first pre-train a cross-modal Transformer using self-supervision on vision and language pairs and subsequently fine-tune on the specific VLN tasks. While these works learn attention between image region and language, Chen \etal \cite{chen2020topological} propose to learn cross-modal attention between language inputs and spatial topological maps (to represent an agent's environment as a graph whose nodes denote places and the edges denote their connectivity). Given the topological map and natural language inputs, a VLN task using the Transformer model bears resemblance to sequence prediction in NLP. Specifically, at each time instance, the cross-modal Transformer predicts a single node of the topological map in the navigation plan. The individual language and map encodings are first processed using uni-modal encoders and later a cross-modal encoder (similar to LXMERT \cite{tan2019lxmert}) is applied to aggregate information across modalities. To denote positions in the map, a learned trajectory position encoding is appended with the map features. Based on this Transformer setup, \cite{chen2020topological} reports a full navigation system that can freely explore the environment and intelligently plan its actions.

\mh{
CLIP \cite{radford2021learning} is a contrastive approach to learn image representations from text, with a learning objective which maximizes similarity of correct text-image pairs embeddings in a large batch size. Specifically, given a batch of $N$ image-text pairs, CLIP learns a multi-modal embedding space, by jointly training an image-encoder and a text-encoder, such that the cosine similarity of the valid $N$ image-text pairs is maximized, while the remaining $N^2-N$ pairs is minimized. The authors consider ResNet-50 \cite{he2016deep} and Vision Transformer (ViT) \cite{dosovitskiy2020image} for encoding images. The modified Transformer model \cite{vaswani2017attention} as in \cite{radford2019language} is employed for encoding text. CLIP is trained on a large corpus of 400 million image-text pairs
and demonstrates excellent zero-shot transfer capabilities. At inference, the names of classes are used as input to the text-encoder, and similarity of the encoded image is computed with all encoded texts (classes) to find the image-text pair with highest match. The CLIP achieves an astounding zero-shot classification accuracy of 75\% on ImageNet, without using an supervision from ImageNet training set. The authors further demonstrate zero-shot transfer capabilities of the CLIP model on 30 different computer vision benchmarks. Note that
CLIP with ResNet took 18 days to train on 592 V100 GPUs while CLIP with ViT took 12 days on 256 V100 GPUs. This highlights the computational cost of CLIP.}

\subsubsection{Single-stream Transformers}\label{Single-stream Transformers}
Different from two-stream networks like ViLBERT \cite{lu2019vilbert} and LXMERT \cite{tan2019lxmert}, VisualBERT  \cite{li2019visualbert} uses a single stack of Transformers to model both the domains (images and text). The input sequence of text (\eg, caption) and the visual features corresponding to the object proposals are fed to the Transformer that automatically discovers relations between the two domains. Notably, VisualBERT architecture is somewhat similar to VideoBERT \cite{sun2019videobert} (explained in Sec.~\ref{sec:video}), but instead of only focusing on cooking videos, VisualBERT evaluates on various visual-linguistic tasks (\eg, VCR, NLVR, VQA, and visual grounding). 
The VisualBERT model first applies task-agnostic pre-training using two objectives (Fig.~\ref{fig:multi-modal-archs}-e). The first objective simply attempts to predict missing text tokens using the image features and remaining textual tokens. The second objective attempts to differentiate between the true and false caption of a given image. After task-agnostic pre-training, the authors propose to perform task-specific pre-training to bridge the domain gap before the final fine-tuning to the downstream task.

Su \etal \cite{su2019vl} propose a multi-modal pre-training approach to learn features that are generalizable to multi-modal downstream tasks such as Visual Commonsense Reasoning and Visual Question Answering. This endeavor requires adequately aligning the visual and linguistic cues so that an effective composite representation is learned. To the end, \cite{su2019vl} builds on the BERT model and inputs both the visual and language features. The language features correspond to the token in the input sentence and the visual features correspond to the region of interest (RoI) from the input image (obtained via a standard Faster R-CNN). 
Specifically, the model is pre-trained on both the visual-lingual dataset (Conceptual Captions \cite{sharma2018conceptual}) as well as the language-only datasets (\eg, Wikipedia). The loss function is identical to BERT, where the model is trained to predict the masked out words or visual ROIs (Fig.~\ref{fig:multi-modal-archs}-f). In contrary to other works such as UNITER \cite{chen2020uniter}, VL-BERT claims that the visual-linguistic matching tasks are not useful during pre-training, which is in contrast to evidence from later efforts \cite{li2020unicoder}. Their results on several multi-modal tasks show their benefit over the language-only pre-training (\eg, in BERT).

Universal Encoder for Vision and Language (Unicoder-VL) \cite{li2020unicoder} learns multi-modal representations using large-scale image-caption pairs. The language and image inputs are fed to a single Transformer model (with multiple successive encoders) to learn joint embeddings. To this end, it uses masked word prediction, masked object classification, and visual-linguistic matching as self-supervision tasks during pre-training (Fig.~\ref{fig:multi-modal-archs}-d). Notably, the visual-linguistic matching is carried out only at the global level (i.e., image-sentence alignment). The model is evaluated on image-text retrieval, zero-shot learning, and visual commonsense reasoning where it performs better than the previous models such as ViLBERT \cite{lu2019vilbert} and VisualBERT \cite{li2019visualbert}. This shows the significance of rich self-supervised tasks and advocates for a unified Transformer architecture to learn multi-modal features in a common framework. 

The Unified Vision-Language Pre-training (VLP) \cite{zhou2020unified} model uses a single Transformer network for both encoding and decoding stages. This stands in contrast to BERT inspired VLP models \cite{sun2019videobert,sun2019learning,su2019vl,li2019visualbert} which use independent encoder and decoder networks.
Joint modeling of encoding and decoding stages allows the Unified VLP model to perform well for both image captioning and visual-question answering tasks, when fine-tuned on these individual tasks. 
The intuition for shared modeling of encoding and decoding stage stems from the need to better share cross-task information during pre-training. The unified model consists of a stack of 12 Transformer blocks, each with a self-attention layer followed by a feed-forward module.
The self-supervised objectives used for pre-training include masked vision-language predictions. Here, the authors explore two variants i.e., bidirectional and sequence-to-sequence prediction of masked works where different context encodings are used for both types of objectives. 
The proposed approach is evaluated on COCO Captions, Flick 30K Captions and VQA 2.0 and obtains encouraging results compared to previous methods on image captioning and VQA \cite{alberti2019fusion}.  

Universal image-text representation (UNITER) \cite{chen2020uniter} performs pre-training on four large-scale visual-linguistic datasets (MS-COCO \cite{lin2014coco}, Visual Genome \cite{krishnavisualgenome}, Conceptual Captions \cite{sharma2018conceptual} and SBU Captions \cite{Ordonez:2011:im2text}). The learned representations transfer well on downstream tasks such as VQA, Multi-modal retrieval, Visual Commonsense reasoning, and NLVR. In order to emphasize on learning the relationships between visual and language domains, \cite{chen2020uniter} specifically designs pre-training tasks to predict masked visual or text region conditioned on the other domain input, and align language and visual inputs on both the global (image-text) and local (word-region) levels (Fig.~\ref{fig:multi-modal-archs}-a). These tasks are beside the conventional masked language modeling task used in BERT and explicitly include fine-grained word-region alignment alongside conditional masking of inputs that were not considered in the earlier works such as VL-BERT \cite{su2019vl}, Visual-BERT \cite{li2019visualbert}, Vilbert \cite{lu2019vilbert} and Unicoder-VL \cite{li2020unicoder}. Common to the other approaches, they adopt the Transformer architecture proposed in BERT that operates on both the visual and language embeddings. In contrast to applying independent Transformers to the language and visual inputs (as in ViLBERT \cite{lu2019vilbert} and LXMERT \cite{tan2019lxmert}), UNITER adopts a single Transformer applied to the textual and image inputs like \cite{li2020unicoder, li2019visualbert, su2019vl}. 

VisualBert \cite{li2019visualbert}, Uniter \cite{chen2020uniter}, VL-BERT \cite{su2019vl}, VilBERT \cite{lu2019vilbert}, and Unicoder-VL \cite{li2020unicoder} models for VLP concatenate image and text features and leave it to the self-attention to automatically discover cross-modal relationships. This can complicate the visual grounding of semantic concepts in an image. To address this problem, Object-Semantics Aligned Pre-Training (Oscar) \cite{li2020oscar} first uses an object detector to obtain object tags (labels), which are then subsequently used as a mechanism to align relevant visual features with the semantic information (Fig.~\ref{fig:multi-modal-archs}-b). The motivation is that the textual content generally pertains to major objects in the image, therefore by explicitly adding those image labels to the input, visual features can be better attended. 
Similar to BERT \cite{devlin2018bert}, Oscar uses a Masked Token Loss for VLP, where different tokens in the textual input and image tags are randomly masked and the model predicts these missing tokens. Further, it also uses a contrastive loss that discriminates between the original and noisy/fake image-tag pairs. The representations thus learned are fine-tuned on VQA, cross-modality retrieval, natural language reasoning, and image captioning tasks to obtain better performances compared to VLP methods that do not use object tags. \sk{The recent VinVL \cite{zhang2021vinvl} approach extends Oscar for the object detection task and learns object instance-centered relationships between visual and language domains using an adapted pretraining scheme. The model is trained on a collection of datasets (MS-COCO, OpenImages, Visual Genome and Objects365) and was demonstrated to precisely relate semantic attributes with the visual information and provided better transferability to the downstream visual comprehension tasks. }


\mh{
\subsubsection{Transformers for Visual Grounding}\label{Transformers for Visual Grounding}
Modulated DETR (MDETR) \cite{kamath2021mdetr} has a CNN and BERT backbone to extract features from image and text inputs, respectively. The visual and text features are then separately linearly projected to a shared space, concatenated and fed to a transformer model (with an architecture similar to DETR) to predict the bounding boxes for objects corresponding to the queries in the grounding text. The model is trained by using a loss which predicts a uniform distribution over all relevant text query tokens specific to the predicted bounding boxes. An additional contrastive loss term ensures correspondence between visual and text embedding.
TransVG \cite{deng2021transvg} is a simple design, where visual and text features are fused together in a transformer module, and the bounding-box corresponding to the query is directly regressed using a learnable token (input to the Transformer module, along-with visual and text features). Referring Transformer \cite{li2021referring} is also a simple one stage design where the text and image features are fused in a Transformer encoder, and the Transformer based decoder then directly regresses bounding boxes or segmentation masks.
Visual Grounding with Transformer \cite{du2021visual} has an encoder-decoder architecture, where visual tokens (features extracted from a pretrained CNN model) and text tokens (parsed through an RNN module) are processed in parallel with two distinct branches in the encoder, with cross-modality attention to generate text-guided visual features. The decoder then computes attention between the text queries and visual features and predicts query-specific bounding boxes.}


\subsection{Video Understanding}\label{sec:video}
Existing approaches for audio-video data analysis generally learn representations on short-length videos (up to a few seconds long), that allow them to encode only short-range dependencies \cite{vaswani2017attention, hochreiter1997long}. Long-range dependency modeling is desirable in various uni-modal and multi-modal learning tasks such as activity recognition \cite{carreira2019short,kay2017kinetics,ging2020coot,seong2019video,wang2020end}. Below, we explain recent approaches that seek to resolve this challenge using the expressivity of Transformer networks. \sk{It is important to note that several of these works \cite{zhou2018end,lee2020parameter,sun2019videobert,girdhar2019video} still employ (pretrained) CNNs to encode image/frame-level features in the videos on top of which Transformers are applied to model wide context. A few exceptions include \cite{neimark2021video,arnab2021vivit,wang2020end,gberta_2021_ICML} which obtain frame-level features also using the ViT based backbones.}

\subsubsection{Joint Video and Language Modeling}
The VideoBERT \cite{sun2019videobert} model leverages Transformer networks and the strength of self-supervised learning to learn effective multi-modal representations. Specifically, VideoBERT uses the prediction of masked visual and linguistic tokens as a pretext task (Fig.~\ref{fig:multi-modal-archs}-c). This allows modeling high-level semantics and long-range temporal dependencies, important for video understanding tasks. Given a video, \cite{sun2019videobert} converts speech to text using off-the-shelf speech recognition systems and applies vector quantization (clustering) to obtain visual features from pre-trained video classification models. The BERT model is then directly applied to these concatenated sequences of language and visual tokens to learn their joint distribution. 
The model can be trained with only-text, video-only, and video+text domains. The resulting model showcases interesting capabilities for cross-modal predictions such as video generation from a given textual input (\eg, captions or cooking recipe) and (video-based) future forecasting. The video+text model uses a visual-linguistic alignment task to learn cross-modality relationships. The definition of this pre-text task is simple, given the latent state of the \texttt{[cls]} token, the task is to predict whether the sentence is temporally aligned with the sequence of visual tokens. Further, the learned representations are shown to be very useful for downstream tasks such as action classification, zero-shot classification, and video captioning. 

Zhou \etal  \cite{zhou2018end} explore Masked Transformers for dense video captioning. This requires generating language descriptions for all events occurring in a video. Existing works on this problem generally operate sequentially i.e., first detect events and then generate captions in separate sub-blocks. \cite{zhou2018end} proposes a unified Transformer network to tackle both tasks jointly, thereby seamlessly integrating the multi-modal tasks of event detection and captioning.  First, a video encoder is used to obtain frame-wise representations followed by two decoder blocks focused on proposing the video events and the captions. Since untrimmed videos are considered, a masking network is used in the captioning decoder to focus on describing a single event proposal. Remarkably, \cite{zhou2018end} was the first approach to target dense video captioning using non-recurrent models and used self-attention in the encoder(applied on CNN derived features) to model broad range context between video frames. Experiments on ActivityNet Captions \cite{krishna2017dense} and YouCookII \cite{zhou2018towards} datasets showed good improvements over previous recurrent network and two-stage based approaches.

\subsubsection{Video Action Recognition}
The traditional CNN based methods in video classification generally perform 3D spatio-temporal processing over limited intervals to understand videos. Neimark \emph{et al.} \cite{neimark2021video} propose Video Transformer Network (VTN) that first obtains frame-wise features using 2D CNN and apply a Transformer encoder (Longformer \cite{beltagy2020longformer}) on top to learn temporal relationships. 
Longformer is an attractive choice to process long sequences (with an arbitrary length $n$) due to its $\mathcal{O}(n)$ complexity. 
The classification token is passed through  a fully connected layer to recognize actions or events. The advantage of using Transformer encoder on top of spatial features is two fold: (a) it allows processing a complete video in a single pass, and (b) considerably improves training and inference efficiency by avoiding the expensive 3D convolutions. 
This makes VTN particularly suitable for modeling long videos where interactions between entities are spread throughout the video length. Their experiments on Kinetics-400 dataset \cite{kay2017kinetics} with various backbones (ResNet \cite{he2016deep}, ViT \cite{vision_transformer} and DeiT \cite{touvron2020deit}) shows competitive performance.


Girdhar \etal \cite{girdhar2019video} use a variant of Transformer architecture to aggregate person-specific contextual cues in a video for action classification and localization. Initially, the model uses a Faster-RCNN \cite{ren2016faster} style processing where a backbone model generates features that are forwarded to the Region Proposal Network to obtain object proposals. Then RoI pooling is applied to generate object-specific features. Multi-head self-attention \cite{vaswani2017attention} is then applied on top of the object features as a cascade of self-attention layers. In each Transformer unit, a particular person feature is treated as the `query' (Q), while the features from the neighboring video clip are used as `key' (K) and `value' (V). The location information is explicitly encoded in the input feature map from which K, V and Q are derived, thus incorporating the positional information in the self-attention. For a given $400$$\times$$400$$\times$$64$ video clip, the key and value tensors are of size $16$$\times$$25$$\times$$25$$\times$$128$, while the query is $128$ dimensional vector.  Although \cite{girdhar2019video} uses only RGB stream, additional modalities like optical flow and audio signal (as in competing works) would further increase the compute complexity. Further, the Transformer model was found to be sub-optimal for action localization, perhaps due to its tendency to incorporate global information. Therefore, it is important to achieve the right trade-off between the global and local context for problems that demand precise delineation (\eg, action localization and segmentation).

\begin{figure}[t]
\centering
    \begin{subfigure}[t]{0.4\textwidth}
      \includegraphics[width=\textwidth]{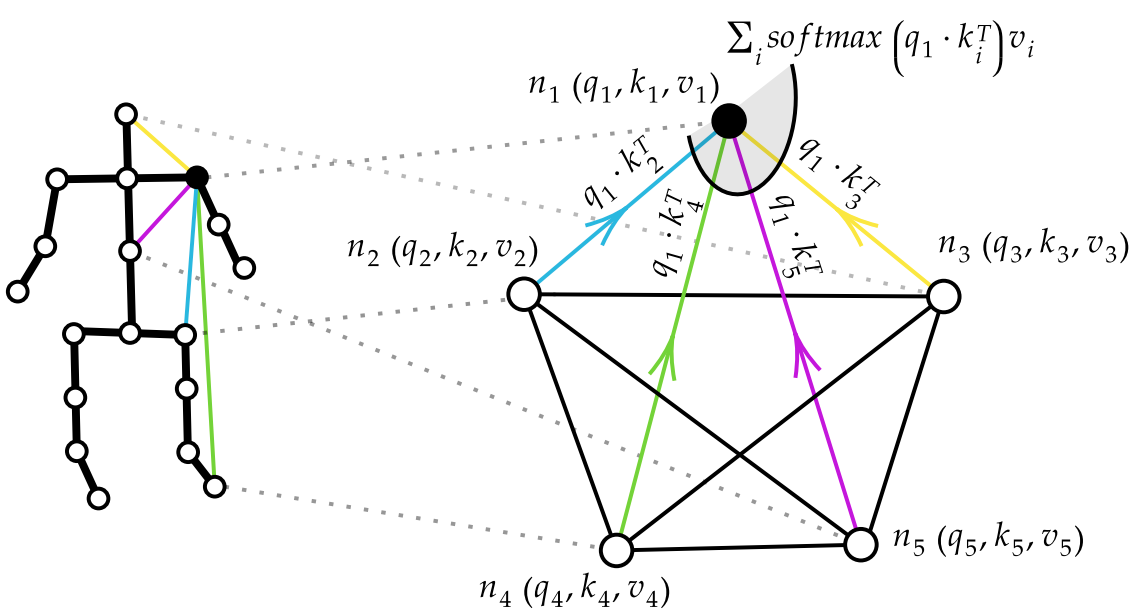}
      \caption{\small Spatial Self-Attention}
      \label{fig:ssa}
    \end{subfigure}
    \begin{subfigure}[t]{0.4\textwidth}
      \includegraphics[width=\textwidth]{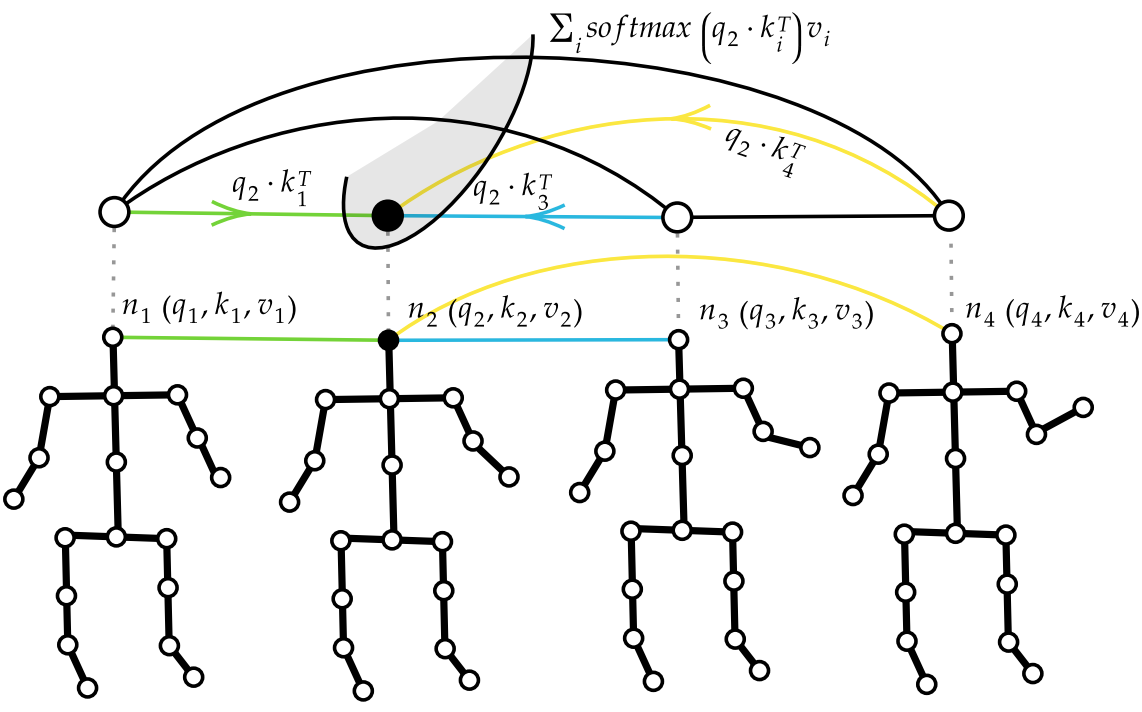}
      \caption{\small Temporal Self-Attention}
      \label{fig:tsa}
    \end{subfigure}
\caption{\small Spatial/Temporal Attention for Skeleton Data Representations. Relationships between body-joints and inter-frame dependencies are modeled using two dedicated self-attention modules. Figure is from \cite{plizzari2020spatial}.}
\end{figure}

Human action recognition based on skeleton representation requires understanding relationships between different joints of a body in a given frame as well as between different frames of a video. Plizzari \etal \cite{plizzari2020spatial} proposed a two-stream Transformer network to model such relationships. They introduced spatial self-attention (SSA) to model relations between different body-joints (Fig.~\ref{fig:ssa})
 while temporal self-attention (TSA) to capture long-range inter-frame dependencies (Fig.~\ref{fig:tsa}).
 They first used a small residual network to extract features from skeleton data and then used SSA and TSA modules to process those feature maps. SSA finds the correlation between each pair of joints independently, while TSA focuses on how features of a certain joint change between frames along the temporal dimension. The purpose of SSA is to discover relationships among the surrounding joints in the same way as the Transformer relates different words in a phrase. On the other hand, TSA finds long-range relations between frames, similar to how relations among phrases are built in NLP. The two streamed model achieves state-of-the-art results on NTU-RGB+D 60 \cite{Shahroudy_2016_NTURGBD} and NTU-RGB+D 120 \cite{Liu_2019_NTURGBD120} datasets.
 
 \mh{
Multiscale Vision Transformers (MViT) \cite{fan2021multiscale} build a feature hierarchy by progressively expanding the channel capacity and reducing the spatio-temporal resolution in videos. They introduce multi-head pooling attention to gradually change the visual resolution in their pyramid structure. TimeSFormer \cite{gberta_2021_ICML} extends ViTs \cite{dosovitskiy2020image} to videos, by considering the video as a sequence of patches extracted from individual frames. To capture spatio-temporal relationships, they propose divided attention i.e., spatial and temporal attentions are separately applied within each block. TimeSFormer demonstrates SoTA performance on action recognition, and can be applied to clips over one minute.  Another notable pure-transformer based model is the Video Vision Transformer (ViViT) \cite{arnab2021vivit}. First, the spatio-temporal tokens are extracted and then efficient factorised versions of self-attention are applied to encode relationships between tokens. However, they require initialization with image-pretrained models to effectively learn the ViT models. There has also been concurrent work on learning sound pretrained models using self-supervised learning with ViTs. An important recent effort is the long-short contrastive learning (LSTCL) framework \cite{wang2021long}, which reconstructs representations from different time-scales (narrow and broad) as auxiliary learning tasks and demonstrates good down-stream performance. 
}

\subsubsection{Video Instance Segmentation}
 The Video Instance Segmentation Transformer (VisTR) \cite{wang2020end} model extends DETR \cite{carion2020end} for video object instance segmentation (VIS) task. Local features are obtained using a backbone CNN on a collection of video frames. An encoder and a decoder Transformer is used similar to DETR to frame the instance segmentation problem as a sequence to sequence prediction task. The input frame-level features are concatenated to form clip representations and the Transformer outputs instance predictions in a order that is consistent across frames. This integrates the object detection and tracking with-in a single unified architecture. The predicted outputs are matched with the ground-truth using bipartitie matching. Similar to Mask R-CNN \cite{he2017mask}, a separate head is used to predict the instance mask based on self-attention and 3D convolutions. The overall results are competitive among the single model approaches on YouTube VIS dataset \cite{yang2019video}, but performs somewhat lower compared to more complex CNN-based models such as MaskProp \cite{bertasius2020classifying}.

\subsection{Transformers in Low-shot Learning}
In the few-shot learning settings, a support set is provided at the inference to adapt to a novel set of categories. Transformer models have been used to learn set-to-set mappings on this support set \cite{ye2020few} or learn the spatial relationships between a given input query and support set samples \cite{doersch2020crosstransformers}. In terms of absolute performance, the patch-wise spatial self-attention between query and support set images excels compared to an image level association learned in \cite{ye2020few}. However, the patch-wise attention computation is computationally expensive. We elaborate on these approaches below.

Doersch \etal \cite{doersch2020crosstransformers} explore the utility of self-supervision and Transformer model for few-shot fine-grained classification, where distribution mismatch exists between training and evaluation phases. 
They develop Cross-Transformer model to relate a given query image with the few-examples available in the support set. To this end, the Transformer finds spatially similar regions in the query and support set images, and the corresponding features are then used to obtain class decisions for the query. The queries in the Transformer architecture are derived from the grid features obtained using the query image. Similarly, grid features from the support images are used to construct keys and values which are in turn used to derive attended outputs. This approach, besides a contrastive self-supervision based training mechanism, leads to the best performance on the challenging Meta-dataset \cite{triantafillou2019meta}.
 
\begin{figure}[t]
    \centering
    \includegraphics[width=1\columnwidth]{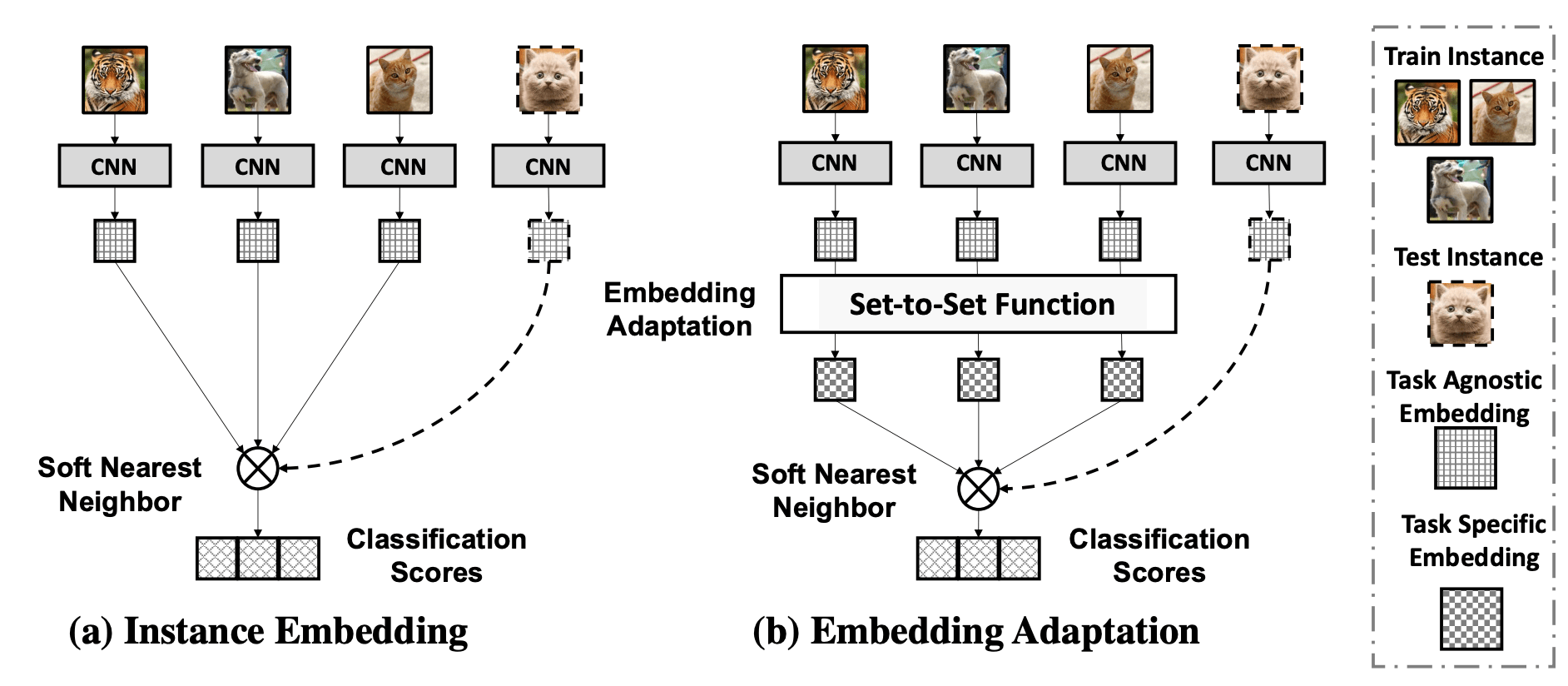}
    \caption{\small An overview of FEAT \cite{ye2020few}. Compared to the conventional instance embedding methods in FSL that keep the embedding function same for all tasks (a), FEAT uses a set-to-set function to adapt the embedding function to each FSL task (b). It evaluates several set-to-set functions and found the Transformer module to be the most suitable choice for FSL. Figure from \cite{ye2020few}.}
    \label{fig:feat}
\end{figure}

Ye \etal \cite{ye2020few} propose to adapt the few-shot embeddings learned on the base classes to the few-shot target classes during inference using a Transformer module. This leads to task-specific embeddings that perform better on the discriminative tasks such as few-shot classification. While many other set-to-set functions are also evaluated, such as Graph convolutional networks \cite{kipf2016semi}, Bidirectional LSTMs \cite{hochreiter1997long} and DeepSets \cite{zaheer2017deep}, the best performance is achieved with the Transformer-based mapping. This is attributed to the better contextualization, task interpolation and extrapolation capability of Transformers and their permutation invariance while maintaining a relatively lower parameter complexity. The Transformer architecture in \cite{ye2020few} follows the standard model \cite{vaswani2017attention}. The embeddings are adapted using a contrastive loss function for preserving discriminative properties (Fig.~\ref{fig:feat}). 
The resulting model achieves strong performance on inductive, transductive, and generalized FSL tasks.  

\mh{
Liu \etal \cite{liu2020universal} learn a multi-head self-attention based module, to integrate the visual representation learned by the models trained on different domains present in the meta-dataset \cite{triantafillou2019meta}. The Universal Representation Transformer (URT) layer dynamically re-weights the representations from different domain-specific backbones, and proves very effective in handling few shot tasks across a variety of data distributions.
}

\subsection{Transformers for Clustering}
Clustering aims to discover structure in the data by grouping similar data points together. It has numerous applications such as data visualization and interpretation, anomaly detection, and open-set categorization. Neural networks have been developed for set prediction problems \cite{zaheer2017deep,edwards2016towards}, however, the setpoints are processed individually which can lose information about inter-point relationships. Recent works employ Transformers that operate on set inputs called the Set Transformers (ST) \cite{lee2019set} for \emph{amortized} clustering. Amortized clustering is a challenging problem that seeks to learn a parametric function that can map an input set of points to their corresponding cluster centers. Lee \etal \cite{lee2019set} propose to learn such a mapping function using a Transformer architecture comprising of multi-head self-attention blocks \cite{vaswani2017attention}.
The Transformer model is permutation invariant by design and allows encoding both pair-wise and higher-order relationships between the input points. 
However, a full Transformer would lead to a  high computational cost of $\mathcal{O}(n^2)$ in each self-attention layer, where $n$ is the number of points in the set. ST reduces this cost to $\mathcal{O}(mn)$ by using an Induced Self-Attention Block that uses a low-rank projection ($H \in \mathbb{R}^m$) to allow operating on large sets. The model was trained to learn optimal parameters that maximize the likelihood of a mixture of Gaussians (MoGs). Thus MoG parameters are estimated by the ST given a set of data points. Beyond amortized clustering, \sk{ST is a generic framework which can handle other set-input problems such as   counting unique elements in an input set, multi-instance learning, set anomaly detection, and 3D point-cloud classification.} More recently,  \cite{lee2019deep} improves \cite{lee2019set} by taking a sequential approach to cluster generation, thereby allowing assignment to a variable number of clusters.

\subsection{Transformers for 3D Analysis}
Given the irregular (variable number of points) and permutation invariant nature of  3D point cloud representations, Transformers provide a promising mechanism to encode rich relationships between 3D data points. To this end, recent works \cite{zhao2020point,guo2020pct} are motivated by the capability of Transformers to learn set-functions. Specifically, \cite{zhao2020point} introduced a Point Transformer which uses vector attention to learn weights for each channel, while \cite{guo2020pct} suggest an alternate design where local 3D structure is explicitly encoded. The non-local nature of Transformers is exploited in \cite{lin2020end} towards an accurate human pose and mesh reconstruction algorithm. We discuss these approaches below.

Self-attention being a set-operator is ideally suited for processing point clouds, a 3D data representation that demands invariance to number of points and their permutations. Zhao \etal~\cite{zhao2020point} propose a point Transformer layer that applies self-attention in the local neighborhood of 3D points. 
 The proposed layer builds on vectorized self-attention network (SAN) \cite{zhao2020exploring} where attention weights are represented with vectors.
 Furthermore, a positional encoding 
 is added both to the attention vector and transformed features (value vectors) to represent location information. The point Transformer layer is sandwiched between two linear layers to create a point Transformer block that is stacked multiple times in the developed network architecture. Their design also included transition down/up blocks to reduce/increase the number of points in the input (in a typical encoding-decoding pipeline style). The resulting architecture shows promising results on the 3D classification and segmentation tasks. 
 
 
 \begin{figure*}[htp]
    \centering
    \includegraphics[width=0.95\textwidth]{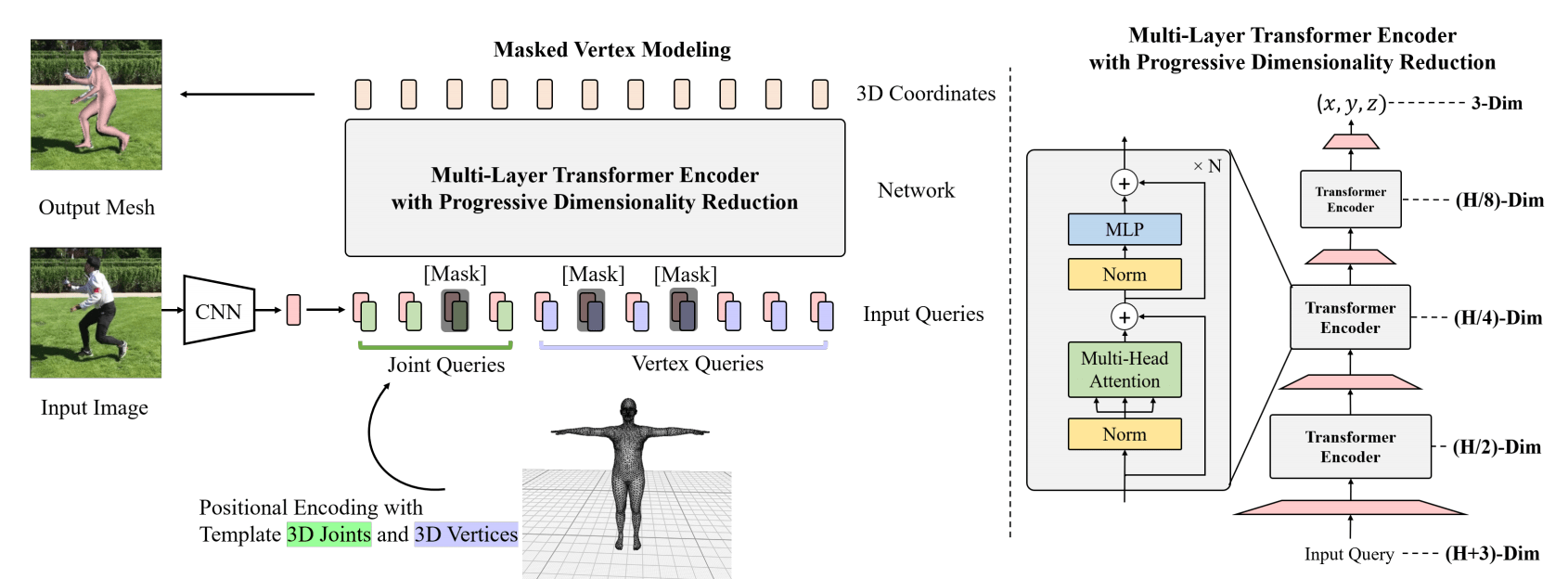}
    \caption{\small Mesh Transformer architecture. The  joint and vertex queries are appended with positional embeddings and passed through multiple self-attention layers to jointly regress 3D coordinates of joints and mesh vertices. Figure is from \cite{lin2020end}. }
    \label{fig:metro}
\end{figure*}

 The Point Cloud Transformer (PCT) \cite{guo2020pct} is a parallel work to \cite{zhao2020point} and motivated by the permutation invariance property of Transformers. However, compared to \cite{zhao2020point}, it is more directly based on the conventional Transformer architecture \cite{vaswani2017attention} and does not involve vector attention. The key modifications include a 3D coordinate-based position encoding, an offset attention module, and a neighbor embedding that encodes local 3D structure in point-clouds. Specifically, the offset attention layer calculates the difference between the self-attended features and the input features using element-wise subtraction. The local neighbor embedding simply finds self-attention relationships among a group of points instead of individual 3D points. Explicitly incorporating local neighbourhood information makes this a more efficient architecture compared to \cite{zhao2020point}. The method shows promising performance on 3D shape classification, normal estimation and segmentation tasks on ModelNet40 \cite{wu20153d} and ShapeNet \cite{shapenet2015} datasets.

The Mesh Transformer (METRO) \cite{lin2020end} model targets 3D human pose and mesh reconstruction from a single 2D image. A key challenge here is to faithfully learn the non-local interactions between body-joints and mesh vertices (\eg, hand and foot). The expressivity of Transformer network is used to jointly model  \emph{vertex to vertex} relationships in a mesh as well as the \emph{vertex to body-joint} relationships. The self-attention mechanism can attend to any combination of vertices in the mesh, thereby encoding non-local relationships. 
The multi-layer Transformer architecture sequentially performs dimensionality reduction to map the 2D image to 3D mesh. Position encoding is performed using the 3D coordinates ($x$,$y$,$z$) of each vertex and each body-joint. Similar to masked language modeling in NLP, METRO uses masked vertex modeling (MVM) which randomly masks some percentage of input queries (see Fig.~\ref{fig:metro}). The Transformer is tasked with regressing all the joints and vertices which helps encode inter-dependencies between them.
METRO obtains state-of-the-art results on human mesh reconstruction on Human3.6M \cite{ionescu2013human3} and 3DPW \cite{von2018recovering} datasets. Since the approach does not depends on a parametric mesh model, it generalizes well to other reconstruction tasks such as 3D hand reconstruction \cite{Freihand2019}. Overall, this is the first effort to employ Transformers for 3D human reconstruction tasks and leads to fairly good results.


\section{\sk{Open Challenges} \& Future Directions}
Despite excellent performance from Transformer models and their interesting salient features (Table \ref{tab:key_features}), there exist several challenges associated with their applicability to practical settings (Table \ref{tab:performance_advantages}). The most important bottlenecks include requirement for large-amounts of training data and associated high computational costs.  There have also been some challenges to visualize and interpret Transformer models. In this section, we provide an overview of these challenges, mention some of the recent efforts to address those limitations and highlight the open research questions.

\subsection{High Computational Cost}
\sk{As discussed in Sec.~\ref{sec:introduction}, a strength of Transformer models is their flexibility to scale to high parametric complexity. While this is a remarkable property that allows training enormous sized models, this results in high training and inference cost (a detailed comparison between CNN and ViTs is shown in Table~\ref{tab:parameters}).} As an example, the BERT \cite{devlin2018bert} basic model (with 109 million parameters) took around 1.89 peta-flop days\footnote{A peta-flop day is a measure of computation and equals to performing $10^{15}$ neural net operations per second for one complete day.} for training, while the latest GPT3 \cite{brown2020language} model (175 billion parameters) took around 3640 peta-flop days for training (a staggering $\sim$1925$\times$ increase). This comes with a huge price tag, \eg, according to one estimate \cite{lambda-gpt3}, GPT3 training might have cost OpenAI 4.6 million USD. Additionally, these large-scale models require aggressive compression (\eg, distillation) to make them feasible for real-world settings. 

\mh{An empirical study on the scalability of Vision Transformers for number of parameters (ranging from five million to two billion), size of the training datasets (ranging from 30 million to three billion training images), and compute budget (1-10000 TPU core-days) is presented in  \cite{zhai2021scaling}. From this study, We can draw the following conclusions (a) scaling up on compute, model and size of training samples improves performance (b) only large models (with more parameters) can benefit from more training data, and the performance of smaller models platueas quickly and can not leverage from additional data. This indicates that large scale models have the capacity to further enhance their representation learning capabilities. However, with the current designs, scaling upon Transformer models is expensive and compute prohibitive, thus necessitating the need for efficient designs.}

\begin{table*}[hbp]
    \centering \small
    \setlength{\tabcolsep}{5pt}
    \scalebox{0.75}{
    \begin{tabular}{x{2.25cm} c p{6cm} x{3cm} x{2cm} x{3cm}} 
    \toprule[0.15em]
      \rowcolor{mygray} \textbf{Task}  &\textbf{Method}& \textbf{Design Highlights} (focus on differences with the standard form)& \textbf{Input Data Type} & \textbf{Label Type}  & \textbf{Loss} \\
      \midrule[0.1em]
      Image Classification & ViT \cite{vision_transformer} & Directly adopted NLP Transformer Encoder for images,  Mechanism to linearly embed image patches with positional embedding suitable for the Encoder. & 2D Image & Class labels & Cross-entropy\\\cmidrule{2-6}
      & DeiT \cite{touvron2020deit}& Transformer as s student while CNN as a teacher, Distillation tokens to produce estimated labels from teacher, Attention between class and distillation tokens. & 2D Image & Class labels & Cross-entropy, Distillation loss based on KL-divergence \\
      \cmidrule{2-6}
      & CLIP \cite{radford2021learning}& Jointly train image and text encoders on image-text pairs, to maximize similarity of valid pairs and minimize otherwise & 2D Images \& texts & Image-text pairs & Symmetric cross-entropy \\
      \midrule
      Object Detection & DETR \cite{carion2020end} & Linear projection layer to reduce CNN feature dimension, Spatial positional embedding added to each multi-head self-attention layer of both encoder and decoder. Object queries (output positional encoding) added to each multi-head self-attention layer of decoder. & 2D Image & Class labels & Hungarian loss based on bipartite matching between predicted and ground truths\\\cmidrule{2-6}
      & D-DETR \cite{zhu2020deformable} & Deformable Transformer consists of deformable attention layers to introduce sparse priors in Transformers, Multi-scale attention module. & 2D Image & Class labels & Hungarian loss \\
      \midrule
       Low Shot Learning & CT \cite{doersch2020crosstransformers} & Self-supervised pretraining,  Query-aligned class prototypes that provide spatial correspondence between the support-set images and query image. & 2D Image & Pretraining without labels and few-shot learning with Class labels & Normalized Cross-entropy \\
       \midrule
       Image Colorization & ColTran \cite{anonymous2021colorization} & Conditional Row/column multi-head attention layers, Progressive multi-scale colorization scheme. & 2D Image & 2D Image &  Negative log-likelihood of the images \\
       \midrule
       Action Recognition & ST-TR \cite{plizzari2020spatial} & Spatial and Temporal self-attention to operates on graph data such as joints in skeletons. & Skeleton & Action Classes & Cross-entropy \\
       \midrule
        Super-resolution & TTSR \cite{yang2020superresolution} & Texture enhancing Transformer module, Relevance embeddings to compute the relevance between the low-resolution and reference image. & 2D Image & 2D Image &  Reconstruction loss, Perceptual loss defined on pretrained VGG19 features. \\
        \midrule
        Multi-Model Learning & Oscar \cite{li2020oscar} & Transformer layer to jointly process triplet representation of image-text [words, tags, features], Masked tokens to represent text data.  & 2D Image & Captions, Class labels,  Object tags & Negative log-likelihood of masked tokens, Contrastive binary cross-entropy \\
        \midrule
        3D Classification/Segmentation & PT \cite{zhao2020point} & Point Transformer block, Transition down block to reduce cardinality of the point set, Transition up for dense prediction tasks.  &  CAD models, 3D object part segmentation &  Object and shape categories & Cross-entropy\\
        \midrule
        3D Mesh Reconstruction & METRO \cite{lin2020end} & Progressive dimensionality reduction across Transformer layers, Positional Encoding with 3D joint and 3D vertex coordinates, Masked vertex/joint modeling. & 2D Image & 3D Mesh + Human Pose &  $L_1$ loss on mesh vertices and joints in 3D and 2D projection.\\
        \midrule
         Vision and Language Navigation & Chen \etal \cite{chen2020topological} & Uni-modal encoders on language and map inputs followed by a cross-modal transformer, Trajectory position encodings in the map encoder. & Instruction text + RGBD panorama + Topological Environment Map & Navigation Plan & Cross-entropy over nodes and \texttt{[stop]} action\\
         \midrule
          Referring Image Segmentation &  CMSA \cite{ye2019cross} & Multimodal feature, Cross-modal self-attention on multiple levels and their fusion using learned gates. & 2D Image + Language expression & Segmentation mask &  Binary cross-entropy loss \\
          \midrule
          Video Classification & Lee \etal \cite{lee2020parameter} & Operates on real-valued audio-visual signals instead of tokens, Contrastive learning for pre-training, End-to-end multimodal transformer learning. & Audio-Visual & Activity labels &  Contrastive InfoNCE loss and Binary cross-entropy \\
    \bottomrule[0.1em]
    \end{tabular}}
    \vspace{0.4em}
    \caption{A summary of key design choices adopted in different variants of transformers for a representative set of computer vision applications. The main changes relate to specific loss function choices, architectural modifications, different position embeddings  and variations in input data modalities. }
    \label{tab:key_features}
\end{table*}

\begin{table*}[hbp]
    \centering\small
    \setlength{\tabcolsep}{4pt}
    \scalebox{0.75}{
    \begin{tabular}{x{1.5cm} p{2cm} p{1.5cm} p{1.5cm} p{1.5cm} p{5cm} p{5cm}}
    \toprule[0.15em]
        \rowcolor{mygray} \textbf{Task }& \textbf{Method}  & \textbf{Metric} & \hspace{-0.3cm} \textbf{Dataset} & \hspace{-0.5cm} \textbf{Performance} &  \hspace{1.2cm} \textbf{Highlights} & \hspace{1.2cm} \textbf{Limitations} \\
        \midrule[0.1em]
         Image Classification  
        &\pb{ViT \cite{vision_transformer} \\ ICLR'21}
        &Top-1 Acc.
        &ImageNet
        &88.55 
        &\textbf{a)} First application of Transformer (global self-attention) directly on image patches, \textbf{b)} Convolution-free network architecture, \textbf{c)} Outperforms CNN models such as ResNet. 
        & \textbf{a)} Requires training on large-scale data \eg,  300-Million images, \textbf{b)} Requires careful transfer learning to the new task, \textbf{c)} Requires large model with 632-Million parameters to achieve SOTA results.  \\\cmidrule{2-7}
        
        & \pb{DeiT \cite{touvron2020deit} \\ arXiv'20}
        & Top-1 Acc. 
        & ImageNet 
        & 83.10 
        & \textbf{a)} Successfully trains Transformer on ImageNet only, \textbf{b)} Introduces attention-based distillation method. \textbf{c)} Produces competitive performance with small (86-Million parameters) Transformers. 
        & \textbf{a)} Requires access to pretrained CNN based teacher model thus performance depends on the quality of the teacher model. \\ \midrule

        & \pb{Swin-T \cite{liu2021swin} \\ arXiv'21}
        & Top-1 Acc. 
        & ImageNet 
        & 84.5 
        & \textbf{a)} Provides a general purpose backbone for different vision tasks e.g., classification, detection and segmentation \textbf{b)} A hierarchical design  using shifted-windows operation.
        & \textbf{a)} Hard to train from scratch on smaller datasets \textbf{b)} Quadratic compute complexity inherent to the self-attention operation. \\ \midrule
        
        Low-Shot Learning 
        & \pb{CT \cite{doersch2020crosstransformers} \\ NeurIPS'20 }
        & Top-1 Acc. 
        & \pb{ImageNet \\ COCO}  
        & \pb{62.25 \\ 60.35}
        & \textbf{a)} Self-supervised pre-training mechanism that does not need manual labels, \textbf{b)} Dynamic inference using Transformer achieving stat-of-the-art results.
        & Proposed algorithm is limited in its capacity to perform on datasets that lack spatial details such as texture. \\ \midrule
        
        
        Object Detection 
        &\pb{ DETR \cite{carion2020end}\\ ECCV'20 }
        & AP 
        & COCO 
        & 44.9 
        & \textbf{a)} Use of Transformer allows end-to-end training pipeline for object detection,\textbf{ b)} Removes the need for hand-crafted post-processing steps. 
        &  \textbf{a)} Performs poorly on small objects, \textbf{b)} Requires long training time to converge. \\ \cmidrule{2-7}
        
        & \pb{ D-DETR \cite{zhu2020deformable} \\ ICLR'21} 
        & AP 
        & COCO 
        & 43.8 
        &  \textbf{ a)} Achieves better performance on small objects than DETR \cite{carion2020end},\textbf{ b)} Faster convergence than  DETR \cite{carion2020end} 
        &  Obtain SOTA results with 52.3 AP but with two stage detector design and test time augmentations. \\\midrule
        Image Colorization 
        & \pb{
        ColTran \cite{anonymous2021colorization}\\ ICLR'21} 
        & FID 
        & ImageNet 
        & 19.71 
        & \textbf{a)} First successful application of Transformer to image colorization, \textbf{b)} Achieves SOTA FID score. 
        & \textbf{a)} Lacks end-to-end training, \textbf{b)} limited to images of size 256$\times$256. \\ \midrule
        
        Action Recognition 
        & \pb{ ST-TR \cite{plizzari2020spatial}\\arXiv'20}
        & Top-1 Acc. 
        &  NTU 60/120 
        & 94.0/84.7 
        & \textbf{a)} Successfully applies Transformer to model relations between body joints both in spatial and temporal domain, \textbf{b)} Achieves SOTA results. 
        & Proposed Transformers do not process joints directly rather operate on features extracted by a CNN, thus the overall model is based on hand-crafted design. \\
        \midrule
        
        Super-Resolution 
        & \pb{ TTSR \cite{yang2020superresolution}\\ CVPR'20} 
        & \pb{ PSNR/ \\ SSIM }
        &\pb{ CUFED5 \\ Sun80 \\ Urban100 \\Manga109}  
        & \pb{ 27.1 / 0.8\\30.0 / 0.81\\ 25.9 / 0.78 \\ 30.1 / 0.91 } 
        & \vspace{-0.7cm} \textbf{a)} Achieves state-of-the-art super-resolution by using attention, \textbf{b)} Novel Transformer inspired architectures that can process multi-scale features. 
        & \vspace{-0.7cm}  \textbf{a)} Proposed Transformer does not process images directly but  features extracted by a convolution based network, \textbf{b)} Model with large number of trainable parameters, and \textbf{c)} Compute intensive.\\
        
        \midrule
        
        Multi-Model Learning & \pb{ViLBERT \cite{lu2019vilbert} \\ NeurIPS'19} 
        & \pb{ Acc./ \\mAP ($R@1$) } 
        & \pb{ VQA \cite{antol2015vqa}/\\ Retrieval\\ \cite{young2014image}} 
        & 70.6/ 58.2  
        & \vspace{-0.5cm} \textbf{a)} Proposed Transformer architecture can combine text and visual information to understand inter-task dependencies, \textbf{b)} Achieves pre-training on unlabelled dataset. 
        & \vspace{-0.5cm} \textbf{a)} Requires large amount of data for pre-training,\textbf{ b)} Requires fine tuning to the new task.\\
        \cmidrule{2-7}
        & \pb{Oscar \cite{li2020oscar} \\ ECCV'20}
        & \pb{ Acc./ \\mAP ($R@1$)} 
        & \pb{ VQA \cite{goyal2017making}/\\ COCO} 
        & \pb{80.37/57.5}
        & \textbf{a)} Exploit novel supervisory signal via object tags to achieve text and image alignment, \textbf{b)} Achieves state-of-the-art results.
        & Requires extra supervision through pre-trained object detectors thus performance is dependent on the quality of object detectors.\\
        \cmidrule{2-7}
        & \pb{ UNITER \cite{chen2020uniter}\\ECCV'20 } 
        & \pb{Acc./\\Avg. ($R@1/5/10$)} 
        &\pb{ VQA \cite{antol2015vqa}/\\ Flickr30K \cite{plummer2015flickr30k}} 
        &  72.47/83.72
        & \vspace{-0.5cm} Learns fine-grained relation alignment between text and images
        & \vspace{-0.5cm} Requires large multi-task datasets for Transformer training which lead to high computational cost. \\
        \midrule
        
        3D Analysis 
        & \pb{ Point Transformer \cite{zhao2020point}\\ arXiv'20}
        & \pb{Top-1 Acc.\\IoU}
        & \pb{ModelNet40 \cite{wu20153d}} 
        &  \pb{92.8\\85.9} 
        & \vspace{-0.5cm} \textbf{a)} Transformer based attention capable to process unordered and unstructured point sets, \textbf{b)} Permutation invariant architecture.
        & \vspace{-0.5cm} \textbf{a)} Only moderate improvements over previous SOTA, 
        \textbf{b)} Large number of trainable parameters around 6$\times$ higher than PointNet++ \cite{qi2017pointnet++}. \\
        \cmidrule{2-7}
        & \pb{ METRO \cite{lin2020end} \\ arXiv'20}
        & \pb{MPJPE\\PA-MPJPE\\MPVE}
        & 3DPW \cite{von2018recovering}
        & \pb{77.1\\ 47.9\\88.2}
        & \vspace{-0.5cm} \textbf{a)} Does not depend on parametric mesh models so easily extendable to different objects, \textbf{b)} Achieves SOTA results using Transformers.
        & \vspace{-0.5cm} Dependent on hand-crafted network design.\\
    \bottomrule[0.1em]
    \end{tabular} }\vspace{0.2cm}
    \caption{A summary of advantages and limitations of different Transformers based methods in different Tasks.  (CT: Cross Transformers, AP: Average Precision, mAP: mean AP,  IoU: Intersection over Union, FID: Fréchet inception distance,  MPJPE: Mean Per Joint Position Error, MPVE: Mean Per Vertex Error).} 
    \label{tab:performance_advantages}
\end{table*}

\begin{table*}[htp]
	\centering \small
		\setlength{\tabcolsep}{6pt}
		\scalebox{0.85}{
		\begin{tabular} {l c c c c}
			\toprule[0.15em]
		\rowcolor{mygray}	Method & \#Param (M) & GFLOPs & Top-1 Acc (\%)  \\
	\midrule[0.1em]
	ResNet18~\cite{he2016deep}$\star$ & 11.7 & 1.8 & 69.8  \\
	EfficientNet-B3~\cite{tan2019efficientnet}$\star$ & 12.0 & 1.8 & 81.6 \\
	DeiT-T~ \cite{touvron2020deit}& 5.7 & 1.3 & 72.2 \\
	T2T-ViT$_t$-7~\cite{yuan2021tokens}& 5.0 & 1.3 & 71.7 \\ 
	LocalViT-T~\cite{li2021localvit} & 5.9  & 1.3  & 74.8 \\
	CrossViT-T~\cite{chen2021crossvit} &  6.9 & 1.6 & 73.4 \\
	PVTv1-T~\cite{wang2021pyramid} & 13.2 & 1.9 &75.1 \\
	ResT-Lite~\cite{zhang2021rest} & 10.5 & 1.4 & 77.2\\
	CaiT-XXX-24~\cite{touvron2021cait} & 12.0 & 2.5 & 77.6 \\
	PVTv2-B1~\cite{wang2021pvtv2} & 13.1 & 2.1 & 78.7 \\
	Lv-ViT-T~\cite{jiang2021all}  & 8.5 & -- & 79.1 \\
	RegionViT-T~\cite{chen2021regionvit} & 13.8 & 2.4 & 80.4\\
	
	\midrule
	ResNet50~\cite{he2016deep}$\star$   &25.6 &4.1 &76.1 \\
	ResNeXt50-32x4d~\cite{xie2017aggregated}$\star$ &25.0 &4.3 &77.6  \\
	RegNetY-4G~\cite{radosavovic2020designing}$\star$ & 21.0 & 4.0 & 80.0 \\
	EfficientNet-B4~\cite{tan2019efficientnet}$\star$ & 19.0 & 4.2 & 82.9 \\
	DeiT-S~\cite{touvron2020deit}  & 22.1 & 4.6 & 79.9 \\
	PVTv1-S~\cite{wang2021pyramid} & 24.5 & 3.8 & 79.8 \\
	LocalViT-S~\cite{li2021localvit} & 22.4 & 4.6 & 80.8 \\
	CrossViT-S~\cite{chen2021crossvit} & 26.7 & 5.6 & 81.0\\
	TNT-S~\cite{han2021transformer}& 23.8 & 5.2 & 81.3 \\
	Swin-T~\cite{liu2021swin}& 29.0 & 4.5 & 81.3 \\
	NesT-T~\cite{zhang2021aggregating} & 17.0 & 5.8 & 81.5 \\
	T2T-ViT$_t$-14~\cite{yuan2021tokens} & 21.5 & 5.2 & 81.5 \\
	CvT-13~\cite{wu2021cvt} & 20.0 & 4.5 & 81.6 \\
	ResT-B~\cite{zhang2021rest} & 30.3 & 4.3 & 81.6\\
	Twins-SVT-S~\cite{chu2021twins} & 24.0 & 2.8 & 81.7 \\
	PVTv2-B2-Li~\cite{wang2021pvtv2} & 22.6 &3.9 & 82.1 \\
	RegionViT-S~\cite{chen2021regionvit} & 30.6 &5.6 & 82.5\\
	Lv-ViT-S~\cite{jiang2021all} & 26.0 & 6.6 & 83.3 \\
	
    \bottomrule[0.15em]
	\end{tabular}}
	\scalebox{0.865}{
	\begin{tabular} {l c c c c}
			\toprule[0.15em]
		\rowcolor{mygray}	Method & \#Param (M) & GFLOPs & Top-1 Acc (\%)  \\
	\midrule[0.1em]
		
	ResNet101~\cite{he2016deep}  $\star$&44.7 & 7.9 &77.4\\
	ResNeXt101-32x4d~\cite{xie2017aggregated}$\star$ & 44.2 & 8.0 &78.8 \\
	RegNetY-8G~\cite{radosavovic2020designing}$\star$ & 39.0 & 8.0 & 81.7 \\
	EfficientNet-B5~\cite{tan2019efficientnet} $\star$& 30.0 & 9.9 & 83.6 \\
	CvT-21~ \cite{wu2021cvt}& 32.0 & 7.1 & 82.5 \\
	CaiT-S-24~\cite{touvron2021cait} & 32.2 &9.4 & 82.7 \\ 
	T2T-ViT$_t$-19~\cite{yuan2021tokens} & 39.0 & 9.8 & 81.4 \\
	PVTv1-M~\cite{wang2021pyramid} & 44.2 & 6.7 & 81.2\\
	PVTv2-B3~\cite{wang2021pvtv2} & 45.2 & 6.9 & 83.2 \\
	NesT-S~\cite{zhang2021aggregating} & 38.0 & 10.4 & 83.3 \\
	
	\midrule
	ResNet152~\cite{he2016deep} $\star$& 60.2 & 11.6 & 78.3 \\
	CaiT-S-36~\cite{touvron2021cait} & 48.0 & 13.9 & 83.3 \\
	T2T-ViT$_t$-24~\cite{yuan2021tokens} & 64.0 & 15.0 & 82.2 \\
	PVTv1-L~\cite{wang2021pyramid}& 61.4 & 9.8 & 81.7 \\
	TNT-B~\cite{han2021transformer} & 66.0 & 14.1 & 82.8 \\
	Swin-S~\cite{liu2021swin} & 50.0 & 8.7 & 83.0 \\
	Twins-SVT-B~\cite{chu2021twins} & 56.0 & 8.3 & 83.2 \\
    RegionViT-B~\cite{chen2021regionvit} & 72.7 & 13.0 & 83.3\\
    PVTv2-B4~\cite{wang2021pvtv2} & 62.6 & 10.1 & 83.6 \\
    
	\midrule
	ResNeXt101-64x4d~\cite{xie2017aggregated} $\star$ & 83.5 & 15.6 & 79.6\\
	RegNetY-16G~\cite{radosavovic2020designing} $\star$ & 84.0 & 16.0 & 82.9\\ 
	EfficientNet-B6~\cite{tan2019efficientnet} $\star$ & 43.0 & 19.0 & 84.0 \\
	NesT-B~\cite{zhang2021aggregating} & 68.0 & 17.9 & 83.8\\
	ViT-B/16~\cite{vision_transformer} & 86.6 & 17.6 & 79.8 \\
	DeiT-B/16~\cite{touvron2020deit} & 86.6 & 17.6 & 81.8 \\
	Swin-B~\cite{liu2021swin} & 88.0 & 15.4 & 83.3 \\
	Twins-SVT-L~\cite{chu2021twins}& 99.2 & 14.8 & 83.7 \\
	PVTv2-B5~\cite{wang2021pvtv2} & 82.0 & 11.8 & 83.8\\
	Lv-ViT-M~\cite{jiang2021all} &  56.0 & 16.0 & 84.1 \\
	\bottomrule[0.15em]
	\end{tabular}}
	\caption{\sk{A Comparative analysis between different vision transformer and CNN models in terms of their parameter complexity and top-1 (\%) accuracy on ImageNet validation set. For a direct comparison, we consider models that are trained on ImageNet from scratch on input of size 224x224. $\star$ denotes pure CNN-based methods.}} 
	\label{tab:parameters}

\end{table*}

In the language domain, recent works focus on reducing the high complexity of Transformer models (basically arising from the self-attention mechanism \cite{vaswani2017attention} where a token's representation is updated by considering all tokens from the previous layer). For example, \cite{child2019generating,beltagy2020longformer} explore selective or sparse attention to previous layer tokens while updating each next layer token.  Linformer \cite{wang2020linformer} reduces complexity of standard self-attention  operation from $\mathcal{O}(n^2)$ to  $\mathcal{O}(n)$ (both in time and memory requirements). The main idea is to show that a low-rank matrix is sufficient to model the self-attention mechanism. The Reformer model 
\cite{kitaev2020reformer} employed locally-sensitive hashing (LSH) to minimize the complexity of self-attention from $\mathcal{O}(n^2)$ to  $\mathcal{O}(n
log(n))$. \sk{In similar pursuit, the recent Lambda Networks propose to model local context as a linear function which helps reduce complexity of self-attention \cite{bello2021lambdanetworks}. These linear function lambdas are applied to the input query to model contextual relationships between pixels.}

Vyas \etal \cite{vyas2020fast} developed an efficient \emph{cluster attention} to deal with large input sequences that approximates the original self-attention. The cluster attention groups queries into clusters and then computes attention between cluster centers (instead of attention between all the queries that leads to quadratic complexity). The main idea is that the queries close in the Euclidean space should have similar attention distributions. With a fixed number of clusters, this intuition helps reduce the quadratic complexity to linear complexity of $\mathcal{O}(nc)$ with respect to the input sequence length $n$ (where $c$ is the number of clusters). We refer interested readers to a survey on efficient Transformers in NLP \cite{tay2020efficient}.

Similar to the NLP domain, computer vision models also suffer from the high computational cost of Transformer models. For example, image generators that are based on sequence-based Transformers (\eg, iGPT) have a high compute cost limiting their applicability to high-resolution inputs. {The time and memory cost of core self-attention operation in Transformers increases quadratically with the number of patches, i.e. $\mathcal{O}(n^2)$, for $n$ image patches (in some applications, e.g., low-level vision, $n=H
\times W$ where $H,W$ denote the height and width of the image). This is a major drawback of existing Transformers that hinders their application to most tasks involving high-resolution (HR) images, such as object detection and segmentation (in high-level vision), and super-resolution, deblurring, denoising, etc. (in low-level vision). Numerous methods have been proposed that make special design choices to perform self-attention more `efficiently', for instance employing pooling/downsampling in self-attention~\cite{wang2021pvtv2,fan2021multiscale, wu2021p2t}, local window-based attention~\cite{liu2021swin,vaswani2021scaling}, axial-attention~\cite{ho2019axial,dong2021cswin}, low-rank projection attention~\cite{wang2020linformer,xiong2021nystr,tay2005synthesizer}, kernelizable attention~\cite{peng2021random,choromanski2020rethinkingperformer}, and similarity-clustering based methods~\cite{kitaev2020reformer,tay2020sparsesinkhorn}. However, almost all of these approaches either come with a trade-off between complexity and accuracy, require special hardware specifications or are still not applicable to very large images. Therefore, there is a pressing need to develop an efficient self-attention mechanism that can be applied to HR images on resource-limited systems without compromising accuracy. It will be interesting to explore how existing models can be extended to high-dimensional cases \eg, using a \emph{multi-scale transformer} design with a somewhat local context modeling.} By inducing inductive biases based on our understanding of the visual learning tasks (e.g., spatial relationships in the local neighbourhood), the high computational cost can be reduced. Similarly, using sparse attention maps modeled with low-rank factorization in the matrices can also help towards reducing the computational cost \cite{neimark2021video}.

\subsection{Large Data Requirements}
Since Transformer architectures do not inherently encode inductive biases (prior knowledge) to deal with visual data, they typically require large amount of training to figure out the underlying modality-specific rules. For example, a CNN has inbuilt translation invariance, weight sharing, and partial scale invariance due to pooling operations or multi-scale processing blocks. However, a Transformer network needs to figure out these image-specific concepts on its own from the training examples. Similarly, relationships between video frames need to be discovered automatically by the self-attention mechanism by looking at a large database of video sequences. This results in longer training times, a significant increase in computational requirements, and large datasets for processing. For example, the ViT \cite{vision_transformer} model requires hundreds of millions of image examples to obtain reasonable performance on the ImageNet benchmark dataset. The question of learning a Transformer in a data-efficient manner is an open research problem and recent works report encouraging steps towards its resolution. For example, DeiT \cite{touvron2020deit} uses a distillation approach to achieve data efficiency while T2T (Tokens-to-Token) ViT \cite{yuan2021tokens} models local structure by combining spatially close tokens together, thus leading to competitive performance when trained only on ImageNet from scratch (without pre-training). \mh{By incorporating CNNs like feature hierarchies in ViTs to effectively capture  local image cues, ViTs (e.g., CCT \cite{hassani2021escaping}, NesT \cite{zhang2021aggregating}) can be trained from scratch even on small-scale datasets (e.g., CIFAR-10).
Another approach to data efficient training of ViTs is proposed in \etal \cite{chen2021vision}. The authors show that by smoothing the local loss surface using sharpness-aware minimizer (SAM) \cite{foret2020sharpness}, ViTs can be trained with simple data augmentation scheme (random crop, and horizontal flip) \cite{szegedy2016rethinking}, instead of employing compute intensive strong data augmentation strategies, and can outperform their counterpart ResNet models.}

\subsection{Vision Tailored Transformer Designs}
We note that most of the existing works focused on vision tasks tend to directly apply NLP Transformer models on computer vision problems. These include architectures designed for image recognition \cite{vision_transformer}, video understanding \cite{sun2019videobert} and especially multi-modal processing \cite{lu2019vilbert}. Although the initial results from these simple applications are quite encouraging and motivate us to look further into the strengths of self-attention and self-supervised learning, current architectures may still remain better tailored for language problems (with a sequence structure) and need further intuitions to make them more efficient for visual inputs. For example, vector attention from \cite{zhao2020exploring} is a nice work in this direction which attempts to specifically tailor self-attention operation for visual inputs via learning channel-wise attentions. Similarly, \cite{he2021transreid} uses a Jigsaw puzzle based self-supervision loss as a parallel branch in the Transformers to improve person re-identification. A recent work \cite{yuan2021tokens} rearranges the spatially close tokens to better model relationships in spatially proximal locations. \sk{Token distillation \cite{touvron2020deit} from pre-trained CNN models has also been used as a remedy to inject domain biases in the representations. One may argue that the architectures like Transformer models should remain generic to be directly applicable across domains, we notice that the high computational and time cost for pre-training such models demands novel design strategies to make their training more affordable on vision problems.}

\mh{
\subsection{Neural Architecture Search for ViTs }
While Nerual Architecuter Search (NAS) has been well explored for CNNs to find an optimized architecture, it is relatively less explored in Transformers (even for language transformers \cite{so2019evolved, wang2020hat}). Chen \etal \cite{chen2021autoformer} propose a one-shot NAS for vision transformers, called AutoFormer. BossNAS \cite{li2021bossnas} searches for a hybrid architecture (CNN and Transformer).} \sk{Another recent effort studies the trade-off between global and local information in Transformers in the context of vision applications \cite{chen2021glit}. It will be insightful to further explore the domain-specific design choices (e.g., the contrasting requirements between language and vision domains) using NAS to design more efficient and light-weight models similar to CNNs \cite{tan2019efficientnet}.}

\subsection{Interpretability of Transformers}

\sk{Through an extensive set of carefully designed experiments, Naseer \etal \cite{naseer2021intriguing} investigate multiple intriguing properties of ViTs in terms of their generalization and robustness. They show that, compared with CNNs, ViTs demonstrate strong robustness against texture changes and severe occlusions, \eg ViTs retain upto 60\% top-1 accuracy on ImageNet once 80\% of the image content is randomly occluded. 
}
Given the strong performance of Transformer architectures, it is interesting and critical to interpret their decisions, \eg, by visualizing relevant regions in an image for a given classification decision. 
The main challenge is that the attention originating in each layer, gets inter-mixed in the subsequent layers in a complex manner, making it difficult to visualize the relative contribution of input tokens towards final predictions. This is an open problem, however, some recent works \cite{voita2019analyzing,abnar2020quantifying,chefer2020transformer} target enhanced interpretability of Transformers and report encouraging results. Attention roll-out and attention flow methods were proposed in \cite{abnar2020quantifying} to estimate the accurate attentions. However, this method functions in an ad-hoc manner and makes simplistic assumptions \eg, input tokens are linearly combined using attention weights across the layers. Chefer \etal~\cite{chefer2020transformer} note that the attention scores obtained directly via the self-attention process (encoding relationships between tokens) or reassignments in \cite{abnar2020quantifying} do not provide an optimal solution. As an alternative, they propose to assign and propagate \emph{relevancy scores} in the Transformer network such that the sum of relevancy is constant throughout the network. Their design can handle both the positive and negative attributions experienced in the self-attention layer. The proposed framework has an added advantage of being able to provide class-specific visualizations. Despite these seminal works, visualizing and interpreting Transformers is an unsolved problem and methods are needed to obtain spatially precise activation-specific visualizations. Further progress in this direction can help in better understanding the Transformer models, diagnosing any erroneous behaviors and biases in the decision process. It can also help us design novel architectures that can help us avoid any biases.

\subsection{Hardware Efficient Designs}
Large-scale Transformer networks can have intensive power and computation requirements, hindering their deployment on edge devices and resource-constrained environments such as internet-of-things (IoT) platforms. Some recent efforts have been reported to compress and accelerate NLP models on embedded systems such as FPGAs \cite{li2020ftrans}. Li \etal \cite{li2020ftrans} used an enhanced block-circulant matrix-based representation to compress NLP models and proposed a new Field Programmable Gate Array (FPGA) architecture design to efficiently manage resources for high throughput and low latency. They could achieve 27x, 3x and 81x improvements in performance (throughput measured in FPS), reduced power consumption, and energy efficiency relative a CPU for RoBERTa model \cite{liu2019roberta}. Towards this goal, \cite{wang2020hat} proposed to design Hardware-Aware Transformers (HAT) using neural architecture search strategies \cite{bender2018understanding, guo2019single, pham2018efficient}. Specifically, a SuperTransformer model is first trained for performance approximation which can estimate a model's performance without fully training it. This model comprises the largest possible model in the search space while sharing weights between common parts. Eventually, an evolutionary search is performed considering the hardware latency constraints to find a suitable SubTransformer model for a target hardware platform (\eg, IoT device, GPU, CPU).
However, such hardware efficient designs are currently lacking for the vision Transformers to enable their seamless deployment in resource-constrained devices. Further, the search cost of the evolutionary algorithms remains significant with the associated impact of CO2 emissions on the environment.

\sk{\subsection{Towards Integrating All Modalities}
Since Transformers provide a unified design to process different modalities, recent efforts also focus on proposing more generic general purpose reasoning systems based on Transformers.
Inspired by the biological systems that can process information from a diverse range of modalities, Perceiver model \cite{jaegle2021perceiver} aims to learn a unified model that can process any given input modality without making domain-specific architectural assumptions. In order to scale to high-dimensional inputs, Perceiver uses an asymmetric cross attention method to distill input information into low-dimensional latent bottleneck features. Once the features are distilled in a compact and fixed-dimensional form, regular Transformer blocks are applied in the latent space.  The original Perceiver model shows performance competitive to ResNets and ViTs on image classification and can process 3D data, audio, images, video or their combinations. However, this model can only generate fixed outputs e.g., class probabilities. A recent improvement called Perceiver IO \cite{jaegle2021perceiver2} aims to learn models with both flexible inputs as well as arbitrary sized outputs. This allows application to problems which demand structured outputs such as natural language tasks and visual comprehension. While these models avoid modality dependent architectural choices, the learning itself still involves modality dependent choices e.g., specific augmentations or positional encodings. An interesting and open future direction is to achieve total modality-agnosticism in the learning pipeline. }

\section{Conclusion}
Attention has played a key role in delivering efficient and accurate computer vision systems, while simultaneously providing insights into the function of deep neural networks. This survey reviews the self-attention approaches and specifically focuses on the Transformer and bi-directional encoding architectures that are built on the principle of self-attention. We first cover fundamental concepts pertaining to self-attention architectures and later provide an in-depth analysis of competing approaches for a broad range of computer vision applications. Specifically, we include state of the art self-attention models for image recognition, object detection, semantic and instance segmentation, video analysis and classification, visual question answering, visual commonsense reasoning, image captioning, vision-language navigation, clustering, few-shot learning, and 3D data analysis. We systematically highlight the key strengths and limitations of the existing methods and particularly elaborate on the important future research directions. With its specific focus on computer vision tasks, this survey provides a unique view of the recent progress in self-attention and Transformer-based methods. We hope this effort will drive further interest in the vision community to leverage the potential of Transformer models and improve on their current limitations \eg, reducing their carbon footprint.



\ifCLASSOPTIONcompsoc
  \section*{Acknowledgments}
\else
  \section*{Acknowledgment}
\fi
{\footnotesize The authors would like to thank Tim Prangemeier (TU Darmstadt), Luowei Zhou (Microsoft Research), Jason Corso (University of Michigan), Pichao Wang (Alibaba Group), Yuqing Wang (Meituan), Alex Meinke (Uni-Tuebingen), Irwan Bello (Google Brain) and Manoj Kumar (Google Brain) for their helpful feedback on the survey. We would also like to thank Mohamed Afham for his help with a figure. }


\bibliographystyle{ieeetr}
\bibliography{eg_bib}





\end{document}